\documentclass{article}

\usepackage{arxiv}
\usepackage{amsmath}
\usepackage{amssymb}
\usepackage{bm}
\usepackage[utf8]{inputenc} % allow utf-8 input
\usepackage[T1]{fontenc}    % use 8-bit T1 fonts
\usepackage{hyperref}       % hyperlinks
\usepackage{url}            % simple URL typesetting
\usepackage{booktabs}       % professional-quality tables
\usepackage{amsfonts}       % blackboard math symbols
\usepackage{nicefrac}       % compact symbols for 1/2, etc.
\usepackage{microtype}      % microtypography
\usepackage{cleveref}       % smart cross-referencing
\usepackage{lipsum}         % Can be removed after putting your text content
\usepackage{graphicx}
\usepackage{subcaption}
\usepackage{natbib}
\usepackage{doi}
\usepackage{float}
\usepackage{needspace}
\usepackage{placeins}
\usepackage{listings}
\usepackage{xcolor}
\usepackage{caption}
\lstdefinestyle{arxivcode}{
    backgroundcolor=\color{gray!5},
    basicstyle=\ttfamily\footnotesize,
    breaklines=true,
    breakatwhitespace=true,
    numbers=left,
    numberstyle=\tiny\color{gray},
    keywordstyle=\color{blue},
    commentstyle=\color{green!60!black},
    stringstyle=\color{purple},
    frame=single,
    framerule=0.5pt,
    framesep=3pt,
    xleftmargin=8pt,
    tabsize=2,
    language=Python,
    showstringspaces=false,
    captionpos=b,
    belowskip=0.5\baselineskip  % 避免代码块后留白过大
}
\crefname{lstlisting}{listing}{listings}
\Crefname{lstlisting}{Listing}{Listings}

\title{Multi-Gate Residuals}

\date{}

\author{%
    \textbf{Zhizhan Zheng}$^{1}$\thanks{Correspondence to: Zhizhan Zheng \texttt{<zhizhanzh@zju.edu.cn>}, Feiyun Zhang \texttt{<zhangfeiyunzfy@gmail.com>}.}~~
    \textbf{Feiyun Zhang}$^{2*}$~~
    \textbf{Shuchun Liu$^{3}$~~}
    \textbf{Tian Xia$^{1}$~~}
    \textbf{Xi Liu$^{1}$~~}
    \textbf{Dasheng Hu$^{1}$~~}
    \textbf{Hongquan Zhou}$^{1}$\\
    $^{1}$Shanghai Yichuang Information Technology Co.,Ltd.~~~~~
    $^{2}$Independent Researcher~~~~~
    $^{3}$Fudan University\\
}

\renewcommand{\shorttitle}{Multi-Gate Residuals}

\hypersetup{
pdftitle={Multi-Gate Residuals},
pdfsubject={cs.CL, cs.AI, cs.LG},
pdfauthor={Zhizhan Zheng, Feiyun Zhang, Shuchun Liu, Tian Xia, Xi Liu, Dasheng Hu, Hongquan Zhou},
}

\begin{document}
\maketitle

\begin{abstract}

While Attention Residuals has shown some effectiveness in addressing the widespread issue of unbounded activation growth across deep residual layers, it inevitably incurs significant communication overhead.
To circumvent this bottleneck, we propose Multi-Gate Residuals (MGR), which stabilizes activation scales without additional communication burden. 
It utilizes a straightforward scoring and gating mechanism to maintain multi-stream context, coupled with Attention Pooling to extract hidden states from the stream states.
Empirical experiments demonstrate that MGR is practical for large-scale training and deployment, offering tangible performance improvements over existing architectures. 

\end{abstract}

% % keywords can be removed
% \keywords{Residual Connections \and  Massive Activations \and Gate Mechanisms}

\section{Introduction}
Residual connections, introduced in ResNet \citep{he_deep_2015}, have become an indispensable component of modern deep neural networks, particularly Transformer \citep{vaswani_attention_2017} architectures. By providing identity-mapping shortcuts, they effectively mitigate vanishing gradients and enable stable training of networks with dozens or even hundreds of layers. 
In large language models, the standard additive residual update under the PreNorm paradigm \citep{xiong_layer_2020}, $x_{l+1} = x_l + \mathcal{F}(\text{LN}(x_l))$, is widely adopted. This formulation preserves training stability by ensuring smooth gradient flow, but it implicitly treats the network depth as a uniform accumulation path of preceding representations. However, as depth increases, this fixed aggregation mechanism exhibits clear limitations: information undergoes progressive dilution across layers, making it difficult for deeper modules to selectively attend to earlier representations. Moreover, the cumulative nature of the residual stream can lead to unbounded magnitude drift, further attenuating the relative influence of individual layer contributions. This behavior is conceptually analogous to the long-range dependency problem in early recurrent neural networks along the sequence dimension, highlighting the need for adaptive information-routing mechanisms in the depth axis.

In recent years, advanced residual connections and depth attention mechanisms have emerged as parallel research directions that improve depth-wise information flow—through channel expansion and data-dependent cross-layer retrieval, respectively. Although these approaches have shown both theoretical and empirical promise, current solutions still face challenges related to architectural complexity, implementation intricacy, and elevated system overhead. Multi-stream topologies enrich connectivity but introduce additional memory-access patterns and numerical sensitivities during large-scale training. Depth-attention architectures, in turn, typically require extra storage and indexing for sequence and depth key-value pairs, specialized fusion kernels, and complex memory layouts to handle cross-layer dependencies, resulting in deployment costs substantially higher than those of standard Transformer blocks. These practical limitations constrain engineering scalability and provide clear room for improvement. 

In this paper, we introduce a novel depth attention structure that substantially simplifies the relatively cumbersome architectures prevalent in the current depth-attention literature while achieving strong model performance, all while retaining the core advantages of advanced residual connections.

\section{Related Work}
Early explorations of depth-wise information flow began with Highway Networks \citep{srivastava_highway_2015}, which introduced gating mechanisms that allow information to traverse multiple layers directly via “highway” paths, thereby alleviating gradient vanishing. Subsequently, DenseNet \citep{huang_densely_2016} proposed dense connections that concatenate each layer’s output to the inputs of all subsequent layers, promoting feature reuse and enhancing gradient propagation while improving parameter efficiency and training stability. These foundational designs established the basis for later optimizations of information flow in Transformer-based architectures.

Building on these ideas, recent advanced residual connection methods have sought to overcome the inherent bottlenecks of uniform depth aggregation by expanding residual pathways, refining cross-layer topologies, or introducing dynamic weighting. A representative line of work is Hyper-Connections (HC) \citep{zhu_hyper-connections_2025} and its variants, which transform the conventional single-stream residual path into multi-stream parallel structures and employ learnable mixing matrices to enable adaptive feature fusion across depths. This channel-broadening approach enriches the network’s macroscopic topology with only modest per-layer FLOPs overhead.
Manifold-Constrained Hyper-Connections (mHC) \citep{xie_mhc_2025} further projects the residual connection space of HC onto a specific manifold, applying manifold constraints to regulate the diversified connectivity patterns introduced by HC.

In parallel, depth attention has emerged as a complementary direction by extending the attention mechanism itself from the sequence dimension to the depth dimension, thereby enabling data-dependent cross-layer retrieval. 
Mixture-of-Depths Attention (MoDA) \citep{zhu_mixture--depths_2026}, for example, allows each attention head to jointly attend to the current-layer sequence key-value (KV) pairs and to depth KV pairs cached from preceding layers within a unified softmax attention framework. This design supports efficient sequence-depth mixing while avoiding the quadratic overhead associated with traditional dense connections. Attention Residuals (AttnRes) \citep{team_attention_2026} reformulates depth aggregation as softmax attention over preceding layer outputs, employing a small set of pseudo-query parameters to achieve content-aware, input-dependent weighting and thereby advancing depth-wise aggregation from linear summation to a fully attentional paradigm.

Although these methods have advanced the state of the art, practical challenges remain. Multi-stream topologies such as those in mHC can introduce additional engineering complexities, while existing depth-attention implementations similarly involve engineering challenges when scaling to large models in distributed training environments. These factors can result in system overheads higher than those of standard Transformer blocks, limiting scalability and deployment in resource-constrained settings. 

\section{Methodology}

Attention Residuals (AttnRes) indeed delivers a substantial architectural improvement in model performance. However, its Full AttnRes variant requires each layer to aggregate the outputs of all preceding layers, a characteristic that inevitably leads to hard-to-overcome communication and memory bottlenecks.
In contrast, the Hyper-Connection (HC) scheme expands the network's hidden state from a single residual stream into 
$n$ parallel hidden vectors to boost performance. Although this constant-factor expansion should theoretically make it more computationally friendly than AttnRes, the introduction of excessively complex operations along the residual paths yields results that are less than ideal in practice, both in terms of performance and efficiency. 
Our approach integrates the key features of both architectural paradigms, effectively combining their complementary strengths to overcome the respective limitations of each, thereby achieving a better balance between efficiency and performance.

\subsection{Our Architecture}
\begin{figure}[htbp]
    \centering
    % 调整宽度为页面宽度的 80%
    % \includegraphics[width=0.8 \textwidth]{images/flow4.png}
    \includegraphics[width=0.8 \textwidth]{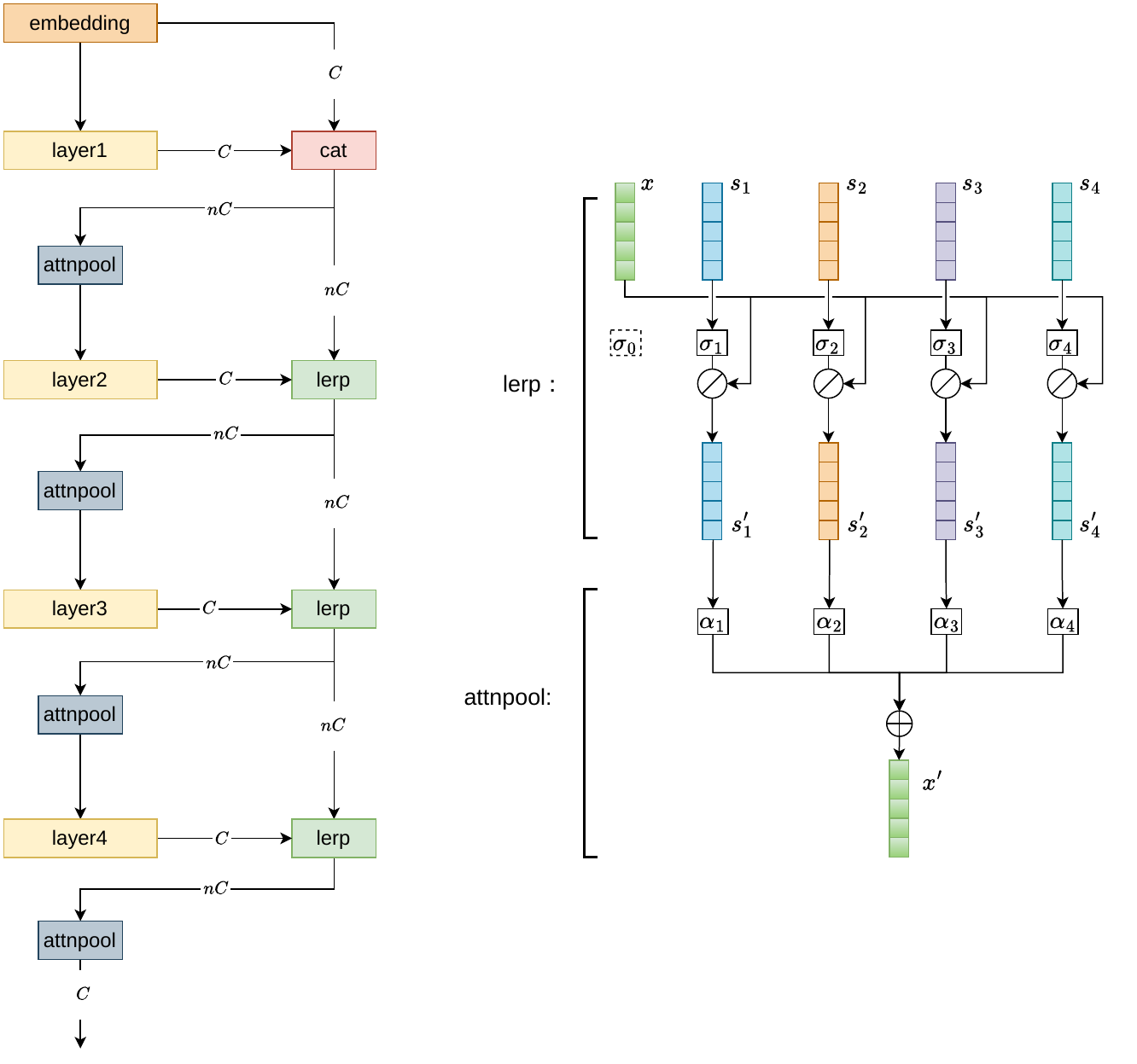}
    \caption{Illustration of Multi Gate Residual Architecture}
    \label{fig:mgr_flow}
\end{figure}

Our architecture, shown in \Cref{fig:mgr_flow}, is extremely simple. The utilization of multi-stream residuals generally necessitates two core structures: aggregators and mixers. 
Their respective roles are to aggregate latent features from the streams for layer processing and to integrate the resulting outputs back into the residual flow width-wise. 
We inherit the depth-wise AttnPool from AttenRes as our aggregator, but configure it to consolidate the cumulative residual stream data relative to the current layer's depth:

\begin{equation}
\begin{aligned}
\alpha_{i\to l} &= \frac{\exp\left( \phi\left(\bm{s}_i, \bm{w}_l^{(\alpha)}\right) \right)}{\sum_{j=1}^{n}\exp\left(\phi\left(\bm{s}_j, \bm{w}_l^{(\alpha)}\right)\right)} \\
\bm{h}_l & = \sum_{i=1}^{n} \alpha_{i\to l} * \bm{s}_i
\end{aligned}
\end{equation}

where the $\bm{w}_l^{(\alpha)}$ as the query is a layer-specific learnable vector in $\mathbb{R}^d$, and adopt the compatibility function as 
$\phi(\bm{s}_i, \bm{w}_l^{(\alpha)}) = \left(\bm{w}_l^{(\alpha)}  \cdot \operatorname{RMSNorm}({\bm{s}_i})\right) / \sqrt{d}$. After normalization $\alpha_{i\to l}$ constitutes a Softmax-weighted distribution that adaptively aggregates residual streams in width.

From the $n$-th layer on, at which point the accumulated residual stream state reaches the predefined threshold $n$, direct concatenation is replaced by gated interpolation, i.e. each stream is mixed with the current layer's output according to its corresponding interpolation coefficient generated by the gating layer.
Two paradigms of gating interfaces are examined as residual stream mixers: an independent gating mechanism using Sigmoid activation, each residual stream is processed by the gating layer with sigmoid function to yield its corresponding interpolation coefficient:
\begin{equation}
\beta_{i\gets l}  = \frac{\exp\left(\varphi\left(\bm{s}_i, \bm{w}_l^{(\beta)}, b_{l, i}^{(\beta)}\right)\right)}{\exp\left(\varphi\left(\bm{s}_i, \bm{w}_l^{(\beta)}, b_{l,i}^{(\beta)}\right)\right) + 1} 
\end{equation}

In the other variant, a competitive multi-gate mechanism is established by applying Softmax to the information flow, ensuring a normalized distribution of weights across the streams:
\begin{equation}
\beta_{i\gets l}  = \frac{\exp\left(\varphi\left(\bm{s}_i, \bm{w}_l^{(\beta)}, b_{l, i}^{(\beta)}\right)\right)}{\sum_{j=1}^{n}\exp\left(\varphi\left(\bm{s}_i, \bm{w}_l^{(\beta)}, b_{l,i}^{(\beta)}\right)\right) + \exp\left(b_{l,0}^{(\beta)}\right)}
\end{equation}

For both cases, we impletended the scoring function $\varphi$ as:
\begin{equation}
\varphi\left(\bm{s}_i, \bm{w}_l^{(\beta)}, b_{l,i}^{(\beta)}\right)  = \frac{\bm{w}_l^{(\beta)} \cdot \operatorname{RMSNorm}(\bm{s}_i)}{\sqrt{d}} + b_{l,i}^{(\beta)}
\label{eq:mixer_scoring}
\end{equation}

\Cref{code:pytorch} provides PyTorch-style pseudocode for both gating mechanisms.
As the behavior in AttnPool, we adopt a learnable vector $\bm{w}_l^{(\beta)}$ in $\mathbb{R}^d$ layerwisely, and the input $\bm{s}_i$ is subjected to RMSNorm before being fed into the gating layer. $\bm{b}_{l,i}^{(\beta)}$ refers to a streamwise gating bias for $1 \leqslant i \leqslant n$. In contrast to independent variant, the competitive one additionally incorporates a forget gate score, denoted by a simple $b_{l,0}^{(\beta)}$ in the formula, to control the actual new input such that the sum of the actual updated weights is less than 1.
Then each of the residual streams could be updated:

\begin{equation}
\bm{s}_i' = (1-\beta_i)\odot  \bm{s}_i + \beta_i \odot  \mathcal{F}_l\left(\bm{h}_l\right)
\label{eq:lerp}
\end{equation}

where $\odot$ denotes element-wise multiplication, with $\beta_i$ broadcast over the feature dimension $D$ to match  $\bm{s}_i$ and layer output $\mathcal{F}_l\left(\bm{h}_l\right)$. Following the insights from Highway Networks \citep{srivastava_highway_2015}, we emphasize the importance of gating bias initialization to facilitate initial carry behavior. The specific initialization schemes are detailed in the following section.

% We distinguish between two distinct gating philosophies: Input-exclusive Gating, where the distribution weights are determined solely by the source states $[x_1, x_2]$, and Integrated State-Candidate Gating, which introduces a post-hoc competition by including the generated candidate $h$ in the scoring pool.""Our results suggest that the Integrated approach, despite its theoretical appeal in balancing relative contributions via Softmax, does not outperform the Input-exclusive baseline. This may be due to the increased complexity of the joint optimization space, where the scoring function must simultaneously normalize heterogeneous features from both the input history and the current transformation

% We examine two paradigms of gating dynamics: Source-only Weighting, where the relative contribution of each component is determined strictly by the input states, and Joint Weighting, which introduces a competitive dependency by incorporating the operator's output $h$ into the scoring function. For single-source scenarios, these weights are realized via Sigmoid activation, whereas for multi-source scenarios, they are normalized using Softmax.
% Preliminary empirical evaluations indicate that incorporating the candidate output into the gating mechanism provides no significant performance gain while increasing computational complexity. Consequently, unless otherwise specified, all proposed architectures discussed hereafter refer to the source-only (input-driven) gating structure.

\subsection{Stabilty Analysis}
\label{sec:stability}

While \textbf{Identity Mapping} facilitates gradient flow by maintaining a unitary gain, it lacks a formal contraction constraint. Specifically, in the residual update $x_{l+1} = x_l + \mathcal{F}_l(x_l)$, the Jacobian matrix of the transformation is given by:
\begin{equation}
\mathbf{J} = \frac{\partial \bm{x}_{l+1}}{\partial \bm{x}_l} = \mathbf{I} + \frac{\partial \mathcal{F}_l}{\partial \bm{x}}
\end{equation}
The spectral radius $\rho(\mathbf{J})$ can easily drift beyond the unit circle (i.e., $\rho(\mathbf{J}) > 1$), particularly when the residual branches exhibit positive eigenvalues. This leads to cumulative variance explosion and numerical instability as network depth increases, manifesting as the "massive activation" phenomenon.

This structural defect prompts a re-evaluation of Highway Networks \citep{srivastava_highway_2015}. Much like ResNet, Highway Networks establish facilitated pathways to reduce gradient attenuation. Crucially, their gated mechanism replaces the rigid unitary gain with a learnable dissipative coefficient 
$(1-g_l)$ via a convex combination:
\begin{equation}
\bm{x}_{l+1} = (1-g_l)\odot \bm{x}_l + g_l \odot \mathcal{F}_l(\bm{x}_l), \quad g_l \in (0, 1)
\end{equation}
introduces a mathematical dissipative structure that allows the model to autonomously regulate its internal energy flow. By constraining the update to a convex sum, the operator norm remains bounded:
\begin{equation}
\|\bm{x}_{l+1}\| \leq (1-g_l)\|\bm{x}_l\| + g_l\|\mathcal{F}_l(\bm{x}_l)\| \leq \max(\|\bm{x}_l\|, \|\mathcal{F}_l(\bm{x}_l)\|)
\end{equation}
providing a mathematically superior basis for numerical stability.

The critical distinction emerges in multi‑layer composition. In a gated deep network, the convex combination enforces a per‑layer ceiling rather than an amplification. For an $L$-layer stack, this bound telescopes into a global guarantee:
\begin{equation}
\|\bm{x}_L\| \leq \max_{l \in \{0,\dots,L-1\}} \bigl\{\|\bm{x}_l\|, \|\mathcal{F}_l(\bm{x}_l)\|\bigr\} \leq \max\Bigl(\|\bm{x}_0\|,\; \max_{l} \|\mathcal{F}_l(\bm{x}_l)\|\Bigr).
\end{equation}
Unlike the unbounded accumulation of a plain ResNet, this layer‑wise maximum ensures that the activation magnitude of the entire deep network is capped by the single most extreme layer, not by the cumulative effect of all layers. The gated mechanism thus transforms depth-induced instability from a multiplicative accumulation problem into a bounded selection problem, yielding a fundamentally more robust architecture for training very deep networks.

The gating mechanisms we propose, whether the independent or the competitive variant, are in essence pure linear mixtures between residual streams $\bm{s}_i$ and the layer output $\mathcal{F}_l(\bm{h}_l)$. As convex combinations, they inherently enjoy non‑expansiveness and strict per‑layer norm ceilings. When combined with the convex‑combination modules already present later in the model (the AttnPool from AttnRes), the entire forward pass is globally guaranteed to keep the feature norms bounded, thereby mitigating the unbounded magnitude growth that otherwise leads to numerical instability and the "massive activation" phenomenon.

\subsection{Gate Bias Initialization}

Consider a gated residual connection with a complementary gate:

\begin{equation}
\bm{x}_{l+1} = (1 - g_l)\odot \bm{x}_l + g_l \odot \mathcal{F}(\bm{x}_l) = \bm{x}_l + g_l \odot \left(\mathcal{F}(\bm{x}_l) - \bm{x}_l\right),
\end{equation}

Assume that the per‑layer increment \(\Delta_l = \mathcal{F}(\bm{x}_l) - \bm{x}_l\) is independent and identically distributed (i.i.d.) across layers with finite variance \(\operatorname{Var}(\Delta)\). According to the principle of variance accumulation in forward signal propagation, the variance of the deep representation \(x_L\) can be approximated as

\begin{equation}
\operatorname{Var}(\bm{x}_L) \approx \operatorname{Var}(\bm{x}_0) + \sum_{l=1}^{L} g_l^2 \operatorname{Var}(\Delta_l)
= \operatorname{Var}(\bm{x}_0) + L \, g_{\mathrm{init}}^2 \operatorname{Var}(\Delta).
\end{equation}

To prevent the signal variance at the end of a deep network from growing linearly with depth \(L\) (i.e., from exhibiting \(O(L)\) explosion), or to prevent the incremental contributions from overwhelming the backbone signal when normalization layers are used, we require the total incremental contribution to remain \(O(1)\). This imposes the following condition on the gate activation:

\begin{equation}
g_{\mathrm{init}}^2 \propto \frac{1}{L} \quad \Longrightarrow \quad g_{\mathrm{init}} \propto \frac{1}{\sqrt{L}}.
\end{equation}

For a sigmoid function biased toward the negative saturation region, \(\sigma(b_l) = \frac{1}{1+e^{-b_g}} \approx e^{b_l}\). Substituting this approximation yields

\begin{equation}
e^{b_l} \propto \frac{1}{\sqrt{L}} \quad \Longrightarrow \quad b_l \approx -\frac{1}{2} \ln(L) + C.
\end{equation}

Thus, the negative magnitude of the gate bias should scale with the natural logarithm of the total number of layers \(L\). Every time the depth doubles, the bias should decrease by approximately \(0.35\) (i.e., \(-\frac{1}{2}\ln 2\)). 

Our initialization principle for gating mixers is as follows. For the weight term $\bm{w}_l^{(\beta)}$ in the mixer scoring function \eqref{eq:mixer_scoring}, we typically initialize it to all zeros, so that the initial gating score is governed exclusively by the bias 
$b_l$, and also ensuring unbiased updating wegiht flow across different streams at the beginning of training.
The bias term $\bm{b}_l$, in both cases it involves an adaptive adjustment based on the actual model depth \(L\) in lerping stage.
Concretely, we take as a reference the bias value that is appropriate for a sigmoid-based gate at a chosen base depth \(L^{(\text{base})}\), and then infer reasonable bias values for other depths and residual-stream count variants from this baseline. 
For example, in our preliminary experiments, we observed that for a model of $n=4$ with a block depth of 12, setting the bias to \(-3\) always yield satisfactory performance. Then taking actual lerping depth \(L^{(\text{base})} = 21 \) and \(b_{\text{init}}^{(\text{base})} = -3\) as the reference, we can derive the following formula for the init bias term:
\begin{equation}
b_{\text{init}} ^{(L)}  = \ln\left(\sqrt{\frac{L}{L^{(\text{base})}}}\left( \exp \left( -{b_{\text{init}}^{(\text{base})}}\right) + 1\right) - n\right)
\end{equation}

where $n$ is the number of residual streams in current layers. 
The $b_{\text{init}} ^{(L)}$ above corresponds to a logit-level offset representing the relative margin by which the forgetting signal is incorporated compared to the original pathway.
So for the competitive mixer layer, we would initialize the forgetting bias by $b_{l, 0}^{(\beta)} = b_{\text{init}} ^{(L)}$ and the streamwise $b_{l, i}^{(\beta)}$ to zero, while for the independent sigmoid mixing layer, $b_{l, i}^{(\beta)} = -b_{\text{init}} ^{(L)}$ should be applied.

\section{Experiment and Analysis}
To evaluate the effectiveness of our proposed multi-gate residuals (MGR), we implement it in language models trained from scratch and compare it with other related architectures (PreNorm \citep{xiong_layer_2020}, mHC-lite \citep{yang_mhc-lite_2026}, AttenRes \citep{team_attention_2026}), measuring their performance difference across different model scales and residual stream numbers (the expansion rate $n$ for HC-styled models). The detailed training configuration for all models is explained in \Cref{sec:hyperp}.
Briefly, we adopt the nanoGPT framework \footnote{https://github.com/karpathy/nanoGPT} due to its ease of reproducibility
and adopt three model scales: S (12 layers, 0.12B parameters),  M (24 layers, 0.35B parameters),  and L (36 layers, 0.77B parameters). 

We use the FineWeb-10BT \citep{penedo_fineweb_2024} dataset which contains \textasciitilde 10 billion tokens.
We closely follow the default training settings of NanoGPT, with a few modifications detailed below. We adopt a default context length of 1024, a global batch size of 512, and a training duration of 20K iterations. This amounts to 10B training tokens which is roughly one epoch over the training set.
We employ the Muon optimizer \citep{jordan2024muon, amsel_polar_2025} for matrix parameters and AdamW \citep{loshchilov_decoupled_2019, kingma_adam:_2014} for all remaining scalar parameters (e.g., RMSNorm weights and linear layer biases, e.t.c.).
We adopt linear learning rate warmup for the first 200 steps, then the cosine decay schedule gradually decays to 10\% of the peak learning rate by the end of training.

\subsection{Overall Performance}
To verify whether our proposed MGR achieves loss improvements comparable to those of competing architectures, we compare the final training and validation losses of models with different residual connection components in \Cref{tab:loss1}. The results clearly demonstrate that MGR outperforms all others across all model scales and residual stream configurations we tested. 
Specifically, at $n=4$ both of independent and competitive MGR slightly outperform the best-performing Full Attnres baseline among the compared models. In contrast, other approaches such as mHC-lite and Block Attnres fall considerably short, while the Pre-Norm baseline lags significantly behind. As expected, this performance advantage becomes more pronounced under the $n=8$ configuration. 

\begin{table}
	\caption{Loss of trained models. We report training and validation loss at the end of training. To mitigate stochastic fluctuations, training loss is computed as a moving average over the last 200 iterations. \textsuperscript{\dag} Our re-implementation of the mHC-lite operator lacks necesssary optimizations to run under identical experimental settings at Scale L.}
	\centering
	\begin{tabular}{lllllll}
		\toprule
		Model Scale & \multicolumn{2}{c}{S}  & \multicolumn{2}{c}{M} & \multicolumn{2}{c}{L}                  \\
		\cmidrule(lr){2-3} \cmidrule(lr){4-5}  \cmidrule(lr){6-7} % 两条短横线，两端自动修剪
		&  Train & Val &  Train & Val &   Train & Val  \\
		\midrule
		Pre Norm                & 2.9280 & 2.9440 & 2.7286 & 2.7314 & 2.6306 & 2.6213 \\
		mHC-lite ($n$=4)        & 2.8992 & 2.9145 & 2.7157 & 2.7082 &  \multicolumn{2}{c}{N/A\textsuperscript{\dag}} \\
		Block AttnRes ($n$=4)   & 2.8951 & 2.9107 & 2.6994 & 2.7009 &  2.6054 & 2.5946 \\
		Block AttnRes ($n$=8)   & 2.8921 & 2.9081 & 2.6952 & 2.6970 &  2.6925 & 2.5921 \\
		Full AttnRes            & 2.8911 & 2.9066 & 2.6930 & 2.6947 &  2.6036 & 2.5920 \\
    \midrule
		Independent MGR ($n$=4) & 2.8887 & 2.9040 & 2.6911 & 2.6929 &  2.6006 & 2.5903 \\
		Competitive MGR ($n$=4) & 2.8889 & 2.9045 & 2.6889 & 2.6911 &  2.6001 & 2.5898 \\
    \midrule
		Independent MGR ($n$=8) & 2.8877 & 2.9034 & 2.6896 & 2.6912 &  2.5994 & 2.5887 \\
		Competitive MGR ($n$=8) & 2.8869 & 2.9020 & 2.6891 & 2.6908 &  2.5966 & 2.5857 \\
		\bottomrule
	\end{tabular}
	\label{tab:loss1}
\end{table}

\begin{figure}[htbp]
  \centering
  \begin{subfigure}[b]{0.45\textwidth}
	% \caption{mean RMS of output}
    % \label{fig:sub1}
    \vspace{0.5em}
    \includegraphics[width=\linewidth]{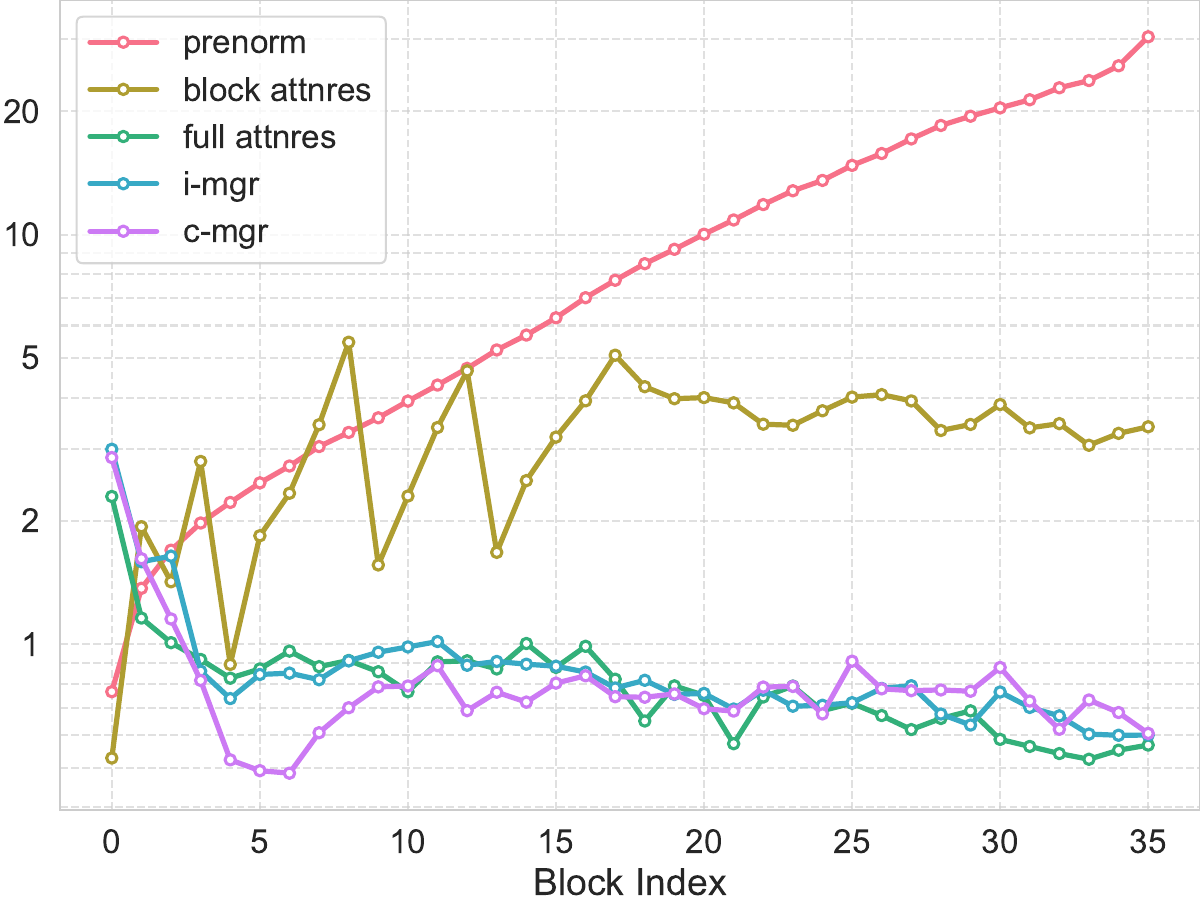}
  \end{subfigure}
  \hfill
  \begin{subfigure}[b]{0.45\textwidth}
    % \caption{per-parameter RMS gradient / $10^{-5}$}
    % \label{fig:sub2}
    \vspace{0.5em}
    \includegraphics[width=\linewidth]{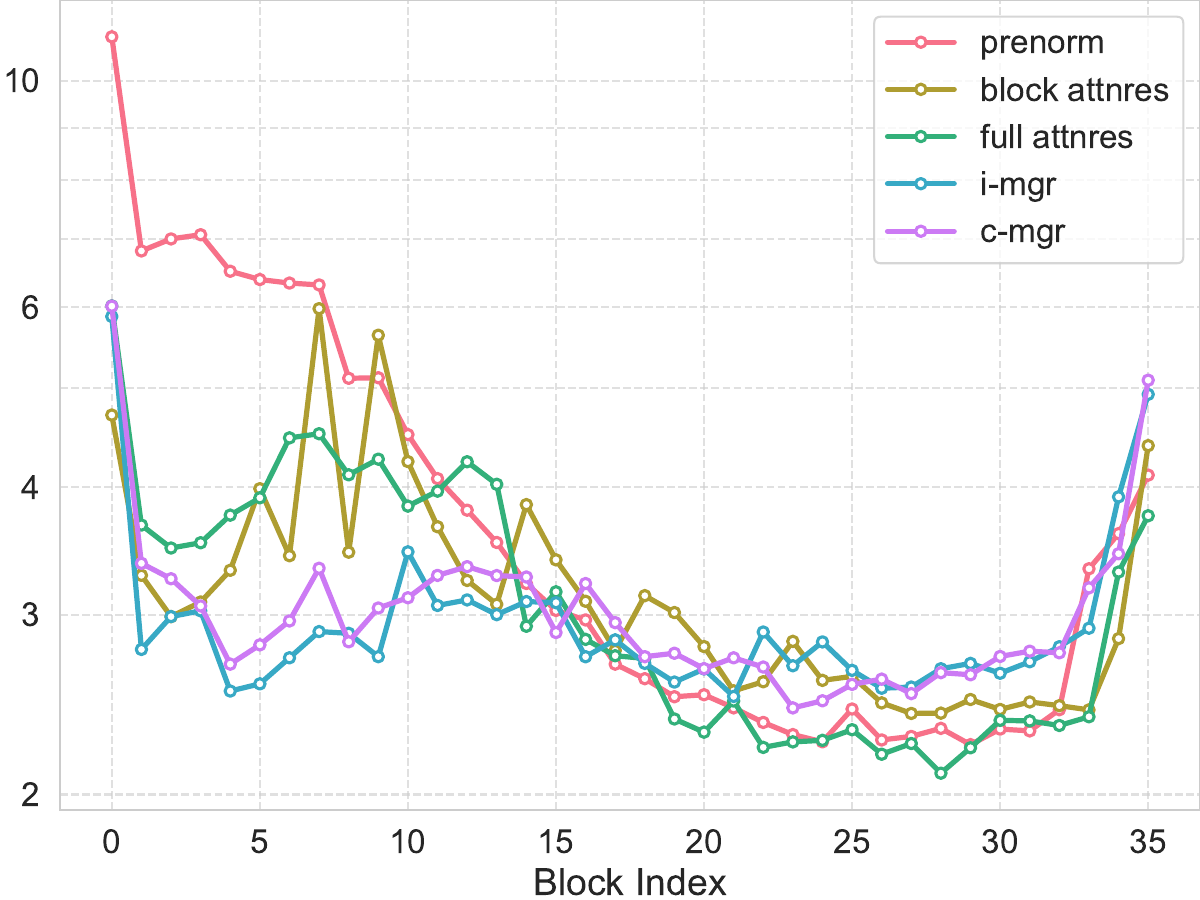}
  \end{subfigure}
  
  \caption{Training dynamics across different architectures. (Left) Output magnitude of each Transformer block at the end of training (measured as average output RMS), (Right) Gradient magnitute of each block (measured as per-parameter RMS gradient, with the y-axis scaled by $10^{-5}$.).}
  \label{fig:train_dynamic}
\end{figure}

We inspected the magnitude of output activations at each Transformer block and the gradient dynamics throughout the network depth af training. \Cref{fig:train_dynamic} summarizes these diagnostics for our proposed methods alongside the main reference architectures.
It is worth highlighting that Block Attnres show substantial variance in per-block output magnitudes under our experimental settings. 
Both independent MGR and competitive MGR substantially mitigate this issue, yielding output behavior comparable to full-attention references.
PreNorm's gradient dilution—characterized by progressively attenuated gradients in deeper layers and disproportionately large gradients in shallow layers—is also markedly ameliorated, with both MGR variants showing substantial improvement, effectively like Full AttnRes.

In deep Pre-Normalized architectures, hidden-state magnitudes tend to expand monotonically with depth. This forces deeper layers to generate disproportionately large outputs from fixed-scale normalized inputs merely to preserve their influence in the residual stream—a phenomenon widely recognized as the PreNorm dilution problem \citep{li_siamesenorm_2026, team_attention_2026}.
Maintaining tight control over layer-wise output magnitudes is therefore essential; any significant depth-dependent amplification strongly indicates the presence of this instability. 
Our approach demonstrably overcomes this challenge.

% \subsection{Growth of Hidden State}

\subsection{Depth-wise Effectiveness Analysis}
To empirically assess the redundancy across layers in our architecture, we utilize two complementary metrics: Performance Drop and Angular Distance, inspired by \citet{gromov_unreasonable_2024, sun_curse_2025}. 
For Angular Distance between layer inputs: 
\begin{equation}
d(\mathbf{x}^l, \mathbf{x}^{l+n}) = \frac{1}{\pi} \arccos\left( \frac{\mathbf{x}^l_T \cdot \mathbf{x}^{l+n}_T}{\|\mathbf{x}^l_T\|_2 \|\mathbf{x}^{l+n}_T\|_2} \right)
\end{equation}
which maps cosine similarity to $[0,1]$, where lower values indicate higher directional consistency, with $0.5$ marking the orthogonality boundary (i.e., uncorrelated features). 

\begin{figure}[htbp]
  \centering
  \begin{subfigure}[b]{0.24\textwidth}
	\caption{}
    \label{fig:heatmap_preln}
    \vspace{0.5em}
    \includegraphics[width=\linewidth]{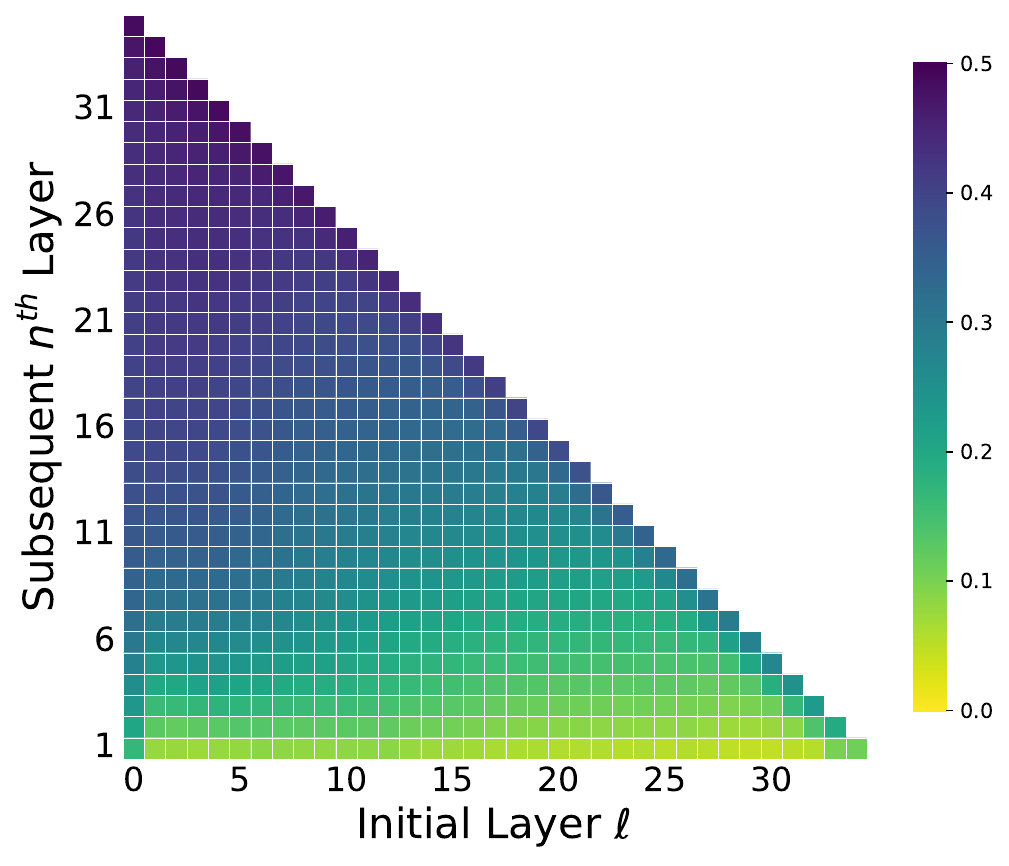}
  \end{subfigure}
  \hfill
  \begin{subfigure}[b]{0.24\textwidth}
	\caption{}
    \label{fig:heatmap_mgr_0}
    \vspace{0.5em}
    \includegraphics[width=\linewidth]{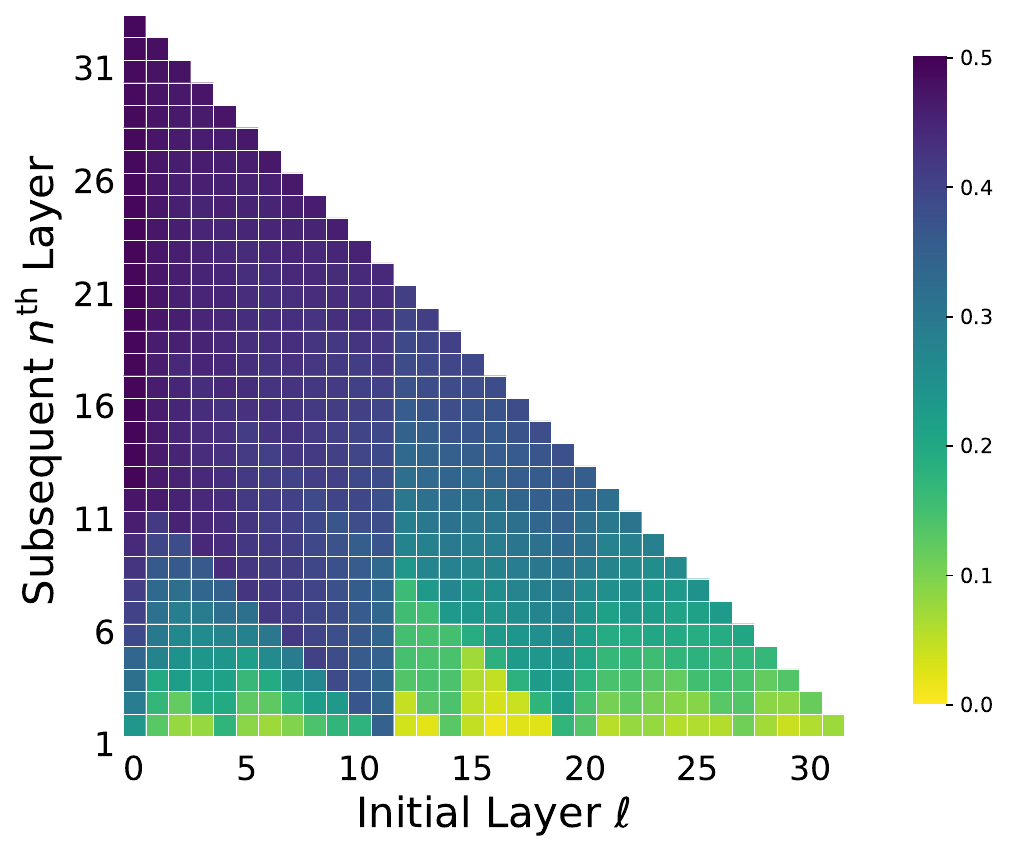}
  \end{subfigure}
  \hfill
  \begin{subfigure}[b]{0.24\textwidth}
	\caption{}
    \label{fig:heatmap_mgr_2}
    \vspace{0.5em}
    \includegraphics[width=\linewidth]{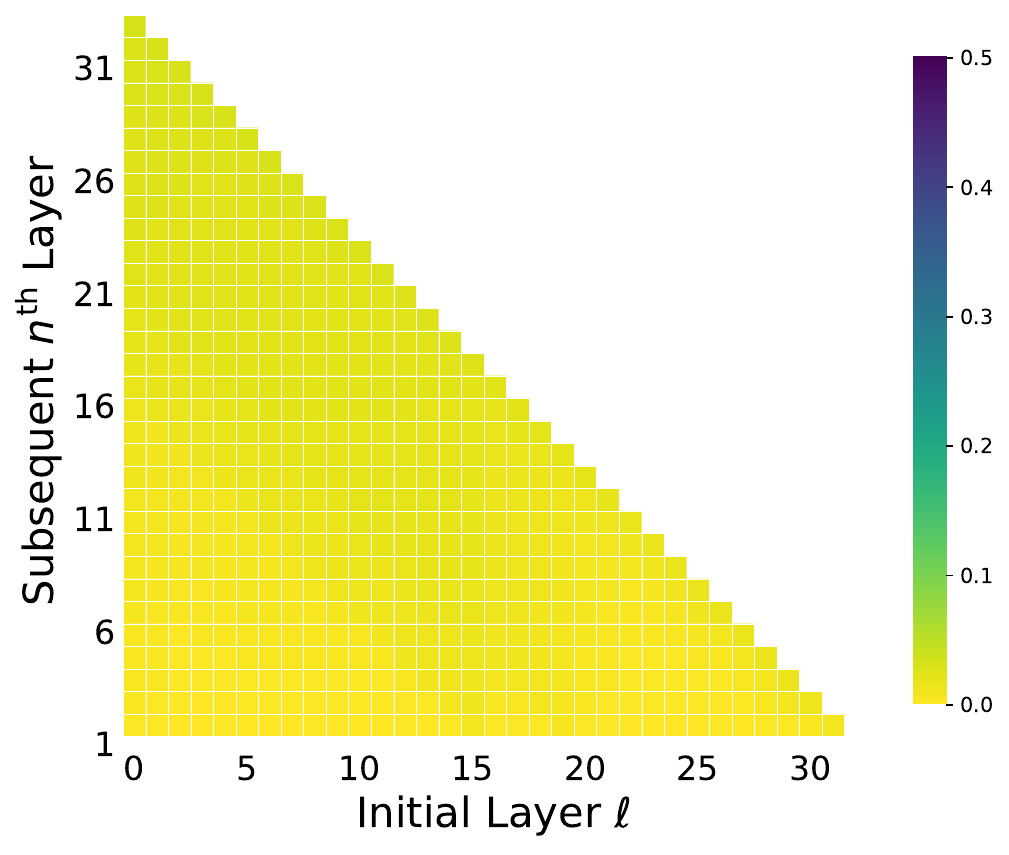}
  \end{subfigure}
  \hfill
  \begin{subfigure}[b]{0.24\textwidth}
	\caption{}
    \label{fig:heatmap_mgr_3}
    \vspace{0.5em}
    \includegraphics[width=\linewidth]{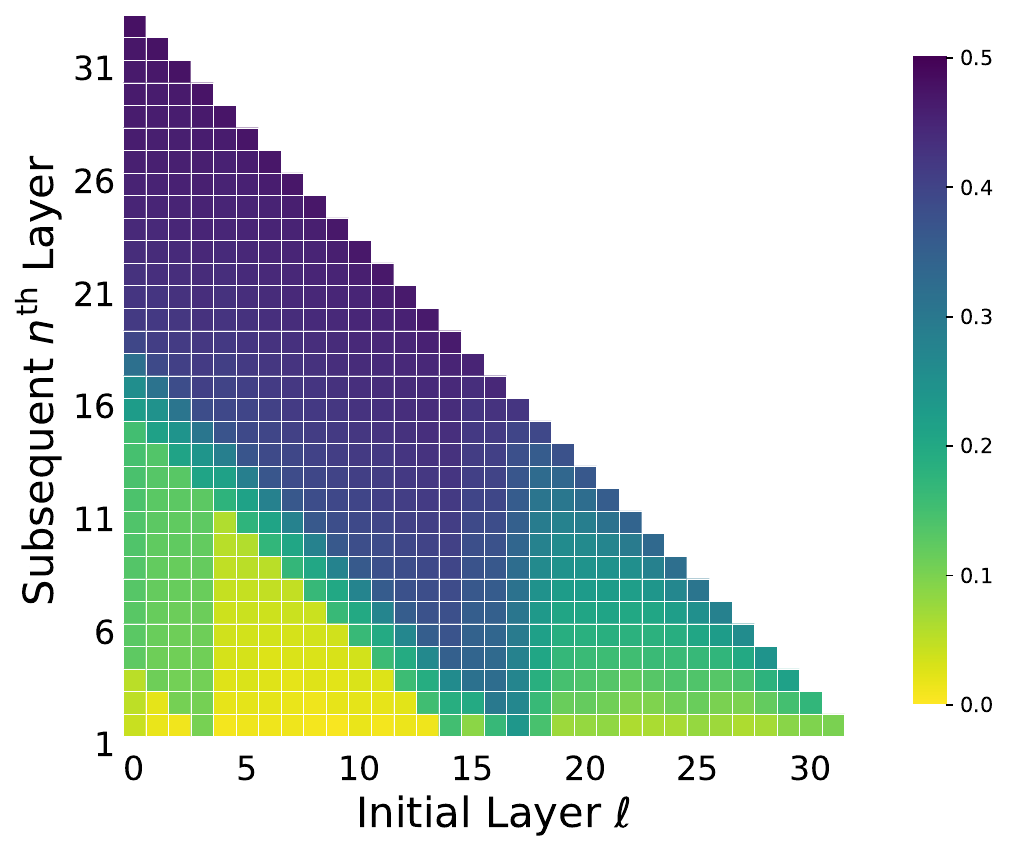}
  \end{subfigure}
  \caption{Angular distance from the initial block $\ell$ (x-axis) and its subsequent $n^{\mathrm{th}}$ block (y-axis). (a) augular distance heatmap from Pre-Norm Architecture, (b-d) augular distance heatmap of the Competitive MGR ($n=8$), showing feature similarity derived from the streams that indexed by 0, 2 and 3, respectively.
  }
  \label{fig:heatmap}
\end{figure}

As illustrated in \Cref{fig:heatmap} the Pre-Norm baseline exhibit decreasing angular distance in deeper layers, indicating highly similar representations. 
In contrast, our MGR architecture reveals heterogeneous similarity patterns along the depth dimension: each stream develops distinct semantic boundary regions where feature divergence sharply increases. Notably, the locations of these semantic discontinuities vary across streams. We hypothesize that aggregating stream-wise features with complementary depth-wise discontinuity patterns enables the model to capture richer hierarchical representations, 
%thereby enhancing discriminative capacity.
thereby preserving multi-granularity semantic information that is otherwise lost to representational homogenization in deep homogeneous architectures.

An interesting empirical pattern emerges in \Cref{fig:heatmap_mgr_2}: 
exactly one stream consistently maintains near-zero angular distances across nearly all layer pairs, implying that their feature representations remain virtually unchanged during forward propagation. While this could reflect a limitation of the current experimental setting—particularly the modest dataset size, which may not provide sufficient signal diversity to differentiate all streams—it may also carry a positive implication. Specifically, the fact that only a subset of streams exhibit such "static" behavior suggests that our choice of stream count is appropriately calibrated to the problem complexity: redundant streams naturally converge to stable representations, while the remaining streams continue to develop depth-wise semantic discontinuities. Future work on larger-scale datasets will help disentangle these two hypotheses and further validate the robustness of our multi-stream design.
The angular distance results for the remaining streams not shown in \Cref{fig:heatmap} are provided in \Cref{sec:ext_layerwise_sim}.

\begin{figure}[htbp]
  \centering
  \begin{subfigure}[b]{0.45\textwidth}
	  % \caption{Pre-Norm Model}
    % \label{fig:sub1}
    \vspace{0.5em}
    \includegraphics[width=\linewidth]{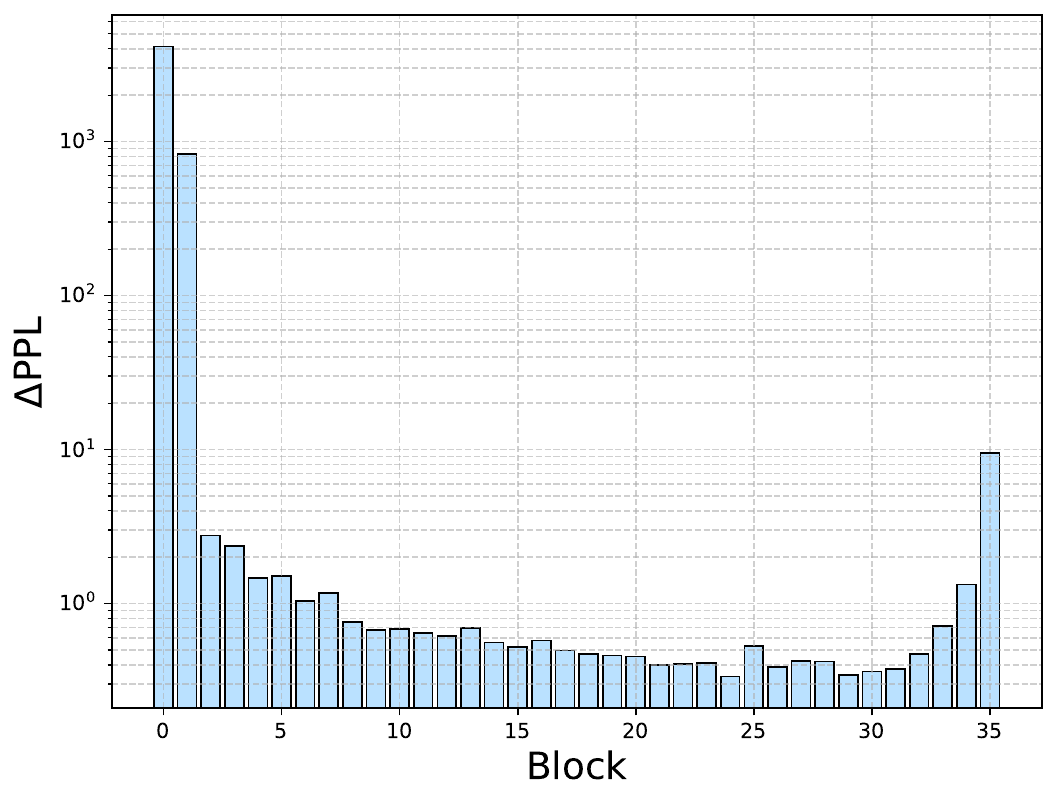}
  \end{subfigure}
  \hfill
  \begin{subfigure}[b]{0.45\textwidth}
    % \caption{competitive MGR ($n=8$)}
    % \label{fig:sub2}
    \vspace{0.5em}
    \includegraphics[width=\linewidth]{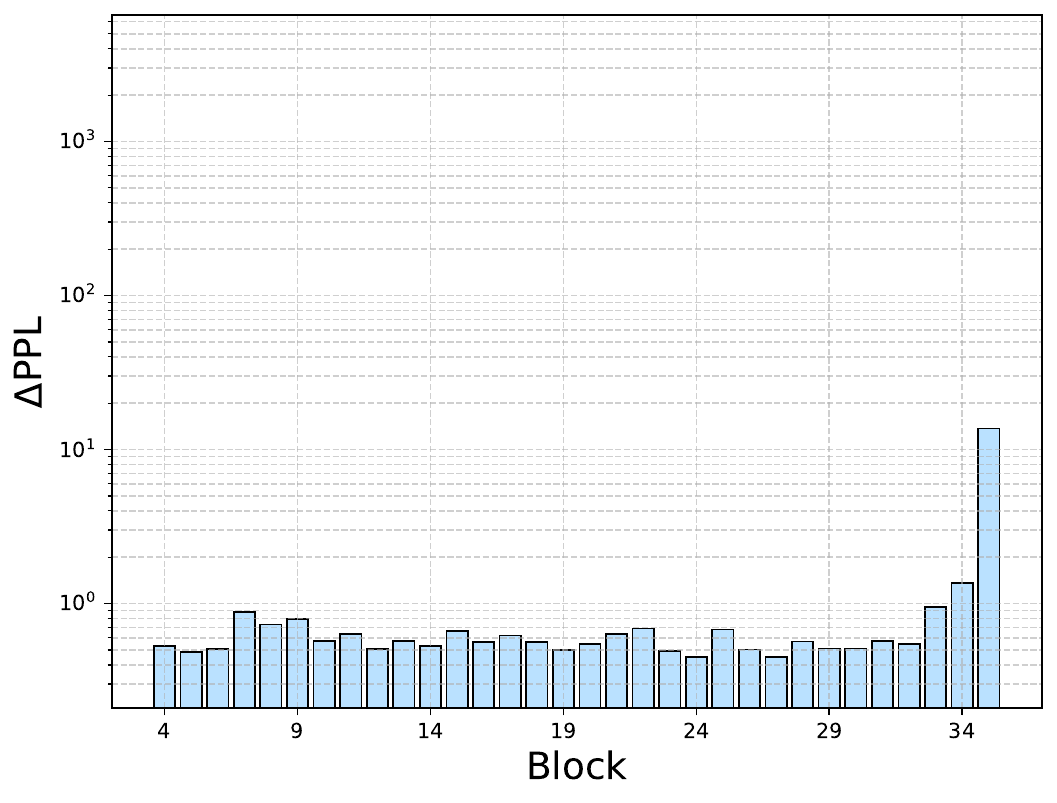}
  \end{subfigure}
  
  \caption{Performance drop ($\Delta \mathrm{PPL}$ in log space) after removing a single block without fine-tuning: (Left) Pre-Norm Model, (Right) competitive MGR ($n=8$). For the competitive MGR profile, we show the block in lerping stage only, earlier layers pruning for MGR is infeasible, as their removal would disrupt the forward pass. }
  \label{fig:bar_pruning_delta_ppl}
\end{figure}

We evaluate both models by assessing the impact of block pruning at different depths, as shown in \Cref{fig:bar_pruning_delta_ppl}. 
Consistent with \citet{sun_curse_2025}, Pre-Norm architectures render early blocks critically important—pruning them induces catastrophic failure—while deeper layers are effectively redundant. 
In sharp contrast, our method achieves a substantially more uniform distribution of inter-layer redundancy. 
The worst-case penalty occurs at final block ($\Delta \mathrm{PPL} = 13.67$) which is structurally expected given its direct adjacency to the language modeling head, yet all pruned variants remain bounded within the same order of magnitude without catastrophic collapse.
% Rather than conforming to the Pre-Norm paradigm of front-loaded structural learning, mid-depth dormancy, and last-minute output assembly by the terminal layers, our method enables deep blocks to sustain effective learning throughout the entire training trajectory and preserves functional engagement even at considerable depth.
This uniform penalty profile indicates that functional engagement is distributed across all depths, with no block entering the semi-dormant regime characteristic of Pre-Norm.

\subsection{Massive Activations}
\begin{figure}[htbp]
  \centering
  \begin{subfigure}[b]{0.32\textwidth}
	\caption{}
    \label{fig:ma_ffn_preln}
    \vspace{0.5em}
    \includegraphics[width=\linewidth]{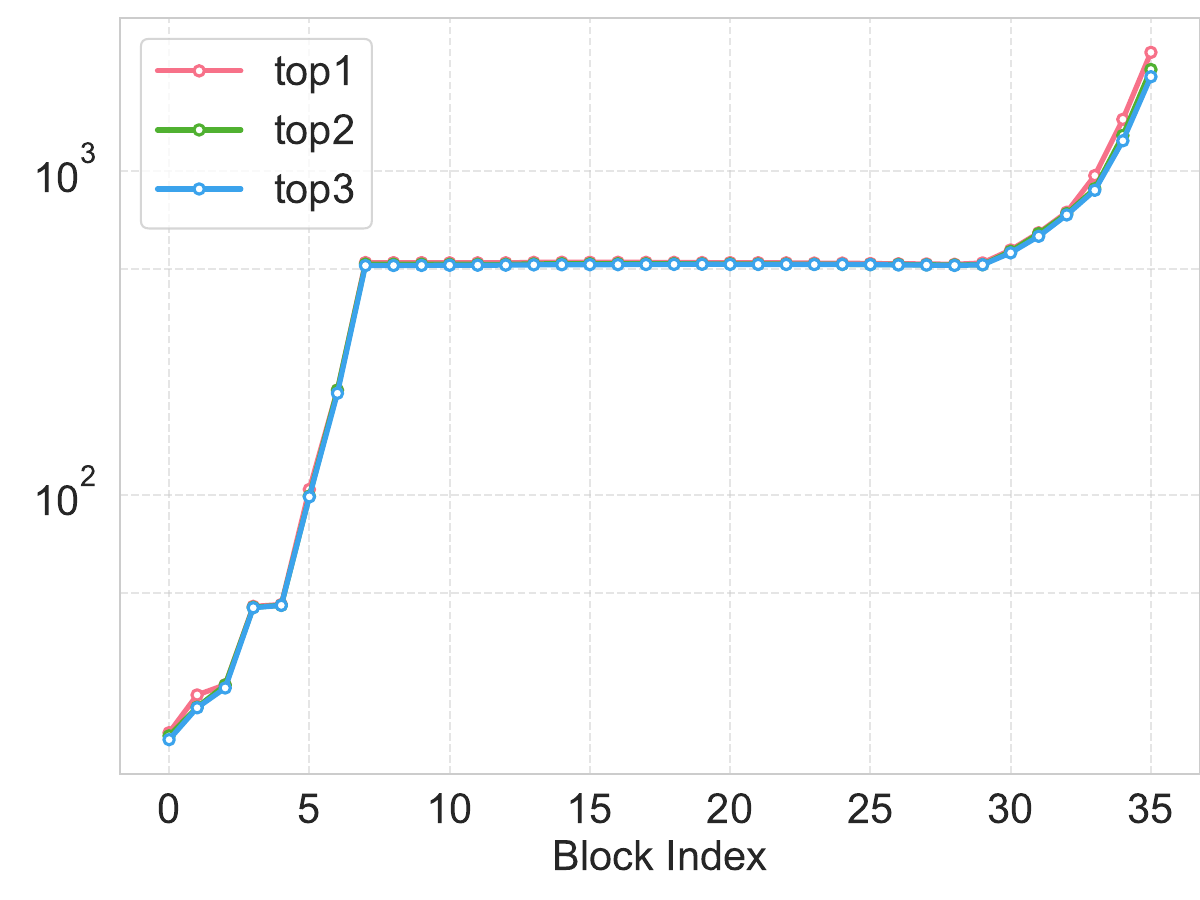}
  \end{subfigure}
  \hfill
  \begin{subfigure}[b]{0.32\textwidth}
	\caption{}
    \label{fig:ma_ffn_fullattnres}
    \vspace{0.5em}
    \includegraphics[width=\linewidth]{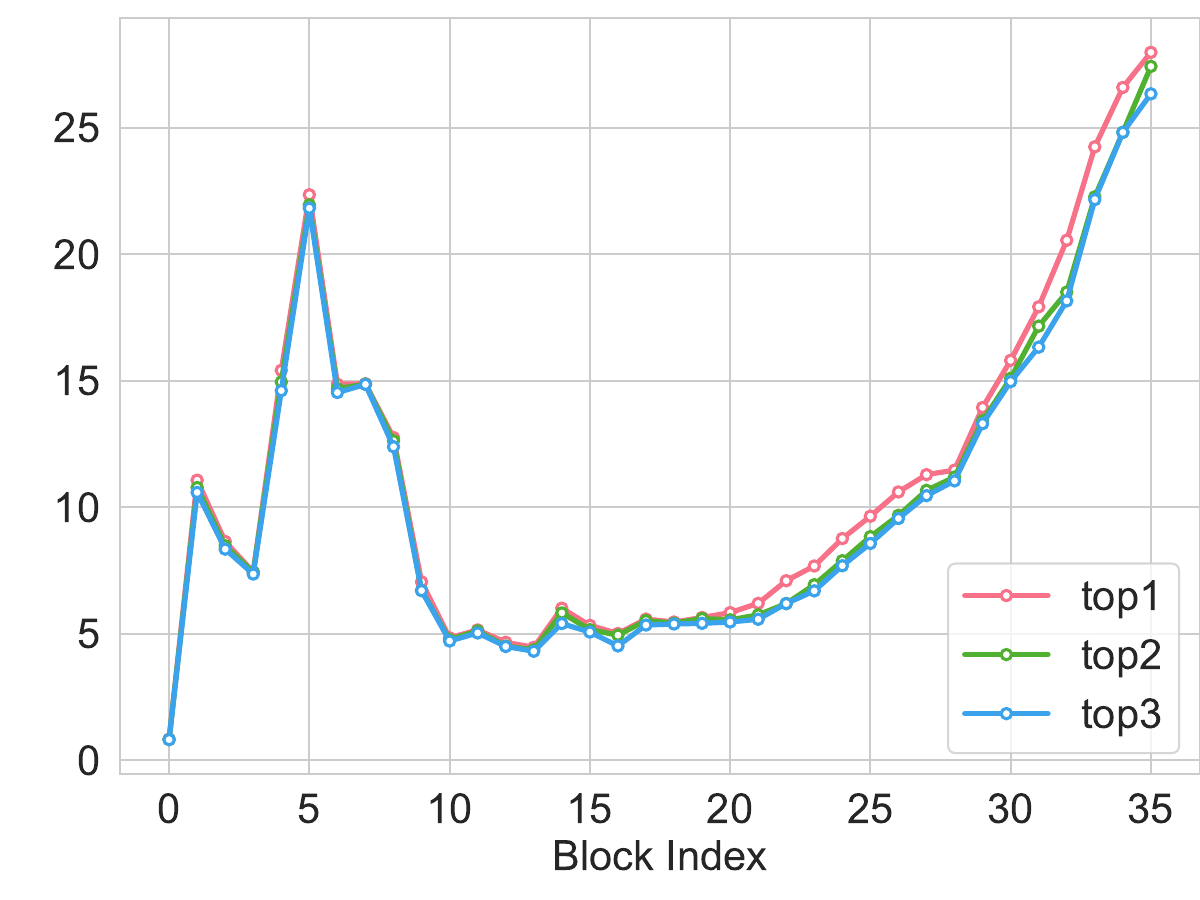}
  \end{subfigure}
  \hfill
  \begin{subfigure}[b]{0.32\textwidth}
	\caption{}
    \label{fig:ma_ffn_cmgr}
    \vspace{0.5em}
    \includegraphics[width=\linewidth]{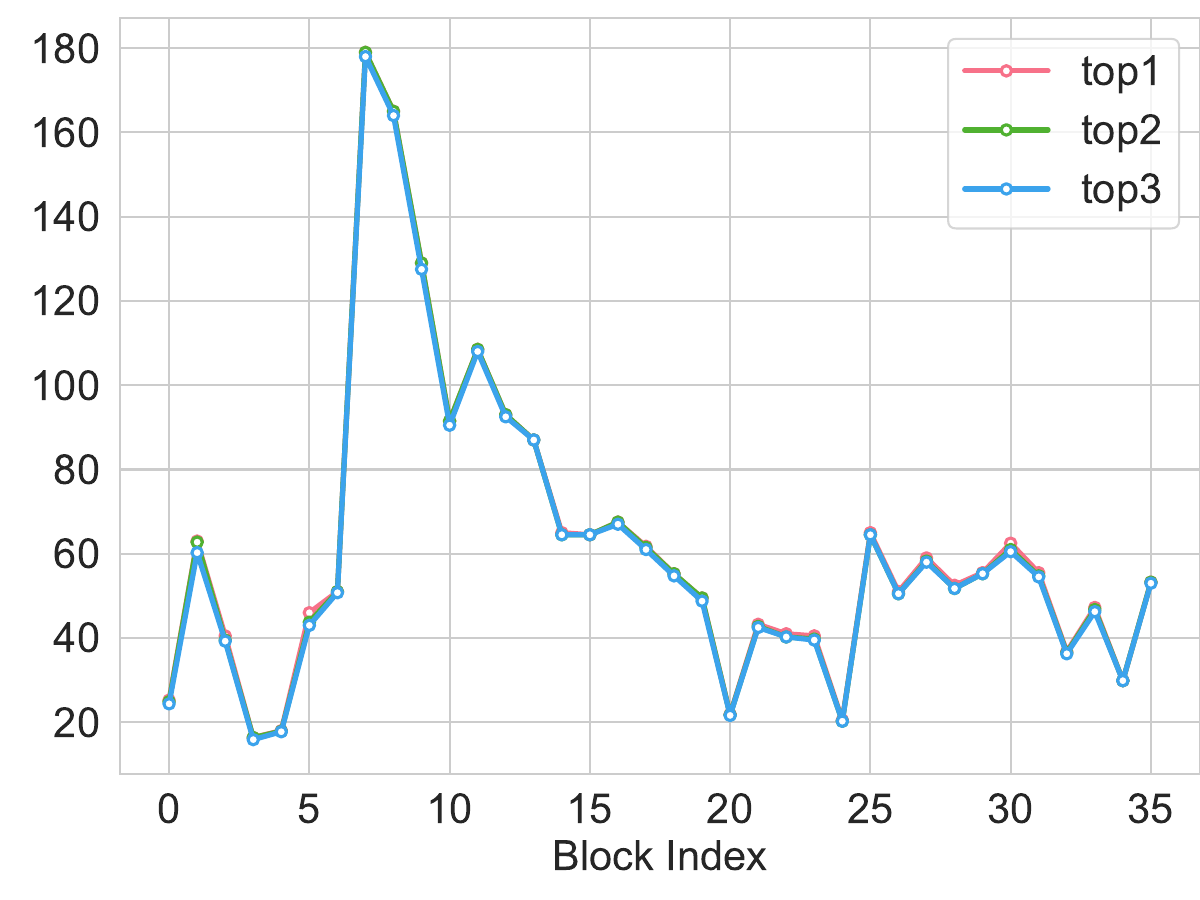}
  \end{subfigure}
  \caption{Comparision of massive activation phenomena across different architectures, with the three largest absolute values of every feedforward layers outputs highlighted for each model. In (a) Pre-Norm baseline: Significant massive activations emerge after the 7th layer. Both in (b) the Full Attnres model, (c) the Competitive model, no significant massive activation were obserbed.
  }
  \label{fig:ma_ffn}
\end{figure}

We analyze the effect of MGR on massive activations by examining the hidden state outputs after every transformer layer.
Specifically, we extract the top three largest activation values from each layer's output and visualize them in \Cref{fig:ma_ffn} (feedforward layers) and \Cref{fig:ma_attn} (attention layers).
These metrics are calculated by running 512 samples of text through the model and collecting all hidden state vectors to evaluate them all at once. 

Observing \Cref{fig:ma_ffn}, the Pre-Norm baseline starts to exhibit massive activation values in early layers, and these values are sustained through subsequent layers, even continuing to rise before the output layer. This late-stage increase differs from \citet{sun_massive_2024}, 
where the activations are observed to drop near the end; nevertheless, this does not affect our main conclusions. 
In \Cref{fig:ma_ffn_cmgr} (our proposed competitive MGR) and \Cref{fig:ma_ffn_fullattnres} (Full AttnRes), the activation magnitudes are considerably smaller, and the sustained plateau of large values seen in  \Cref{fig:ma_ffn_preln} is absent. While MGR yields slightly larger activation magnitudes than AttnRes in the FFN outputs, it achieves smaller magnitudes in the attention outputs (\Cref{fig:ma_attn}); the overall magnitude levels of the two methods are similar.

\begin{figure}[htbp]
  \centering
  \begin{subfigure}[b]{0.45\textwidth}
	%\caption{}
    %\label{fig:ma_stream_ffn_cmgr}
    \vspace{0.5em}
    \includegraphics[width=\linewidth]{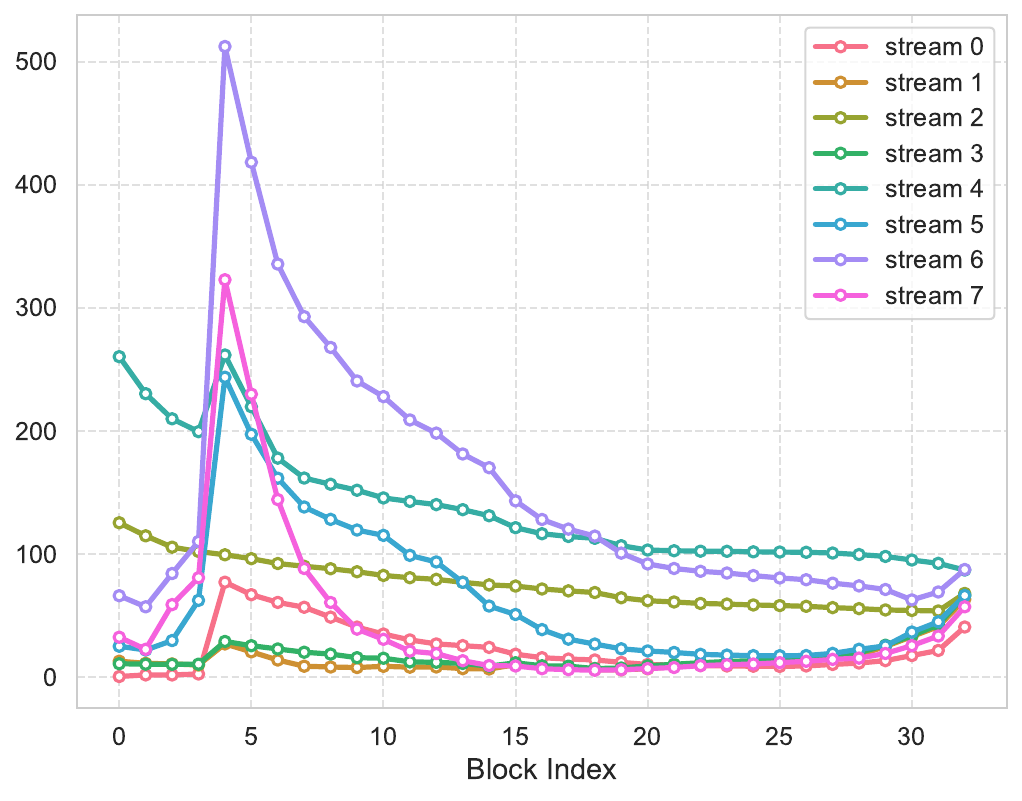}
  \end{subfigure}
  \hfill
  \begin{subfigure}[b]{0.45\textwidth}
	%\caption{}
    %\label{fig:ma_stream_ffn_imgr}
    \vspace{0.5em}
    \includegraphics[width=\linewidth]{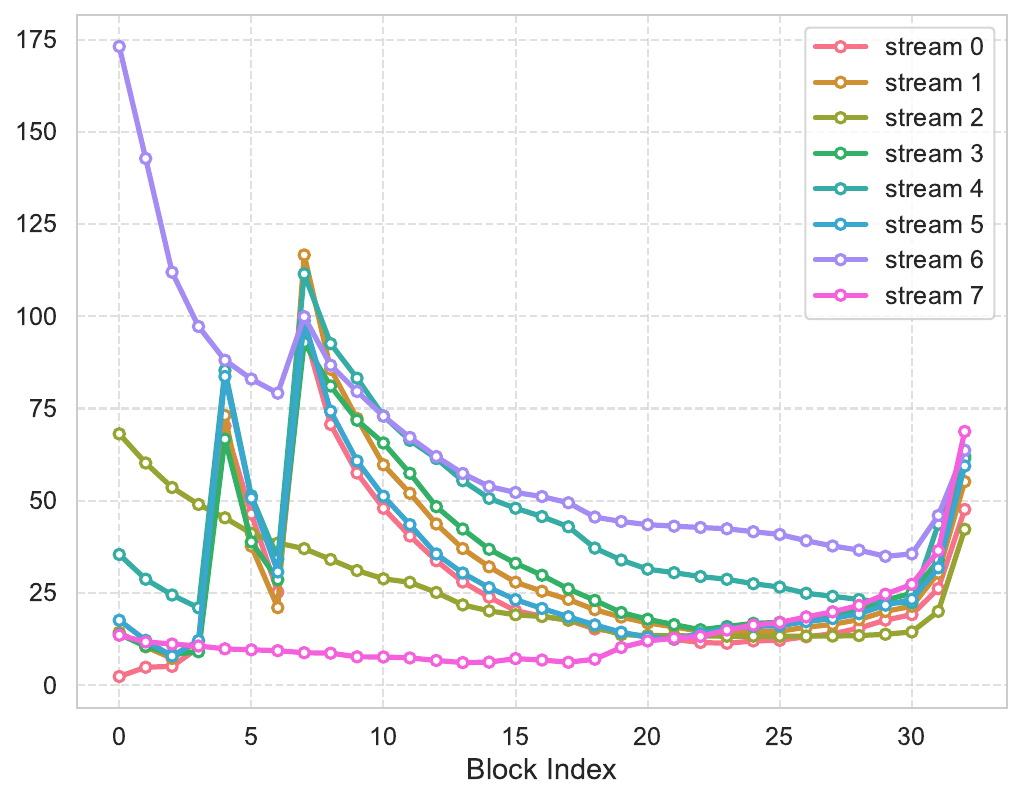}
  \end{subfigure}
  \caption{Maximum absolute value across all output streams for each feedforward layer. (Left): competitive MGR, Right: independent MGR.}
  \label{fig:ma_stream_ffn_mgr}
\end{figure}

We further examined the maximum absolute values of each output stream of competive MGR and independent MGR, as shown in \Cref{fig:ma_stream_ffn_mgr} and \Cref{fig:ma_stream_attn_mgr} (streams merging from attendtion layers and feedforward layers, respectively). For both variants of MGR, the stream outputs remain similarly stable.
As expected, the peak positions of the certain streams coincide with those of the hidden state outputs, this is unsurprising because the hidden state is derived from the stream outputs via attention pooling, after which the stream output magnitudes are rapidly attenuated in the subsequent forward computation.
Across the entire network, the stream magnitudes remain within a reasonable and stable range, which aligns well with our stability analysis in \Cref{sec:stability}.

\subsection{Practical Efficiency Strategies}
\label{sec:practical_efficiency}

MGR's stream-based formulation naturally sidesteps inter-device communication overhead, as long as we choose an appropriate $n$, our optimization focus shifts entirely to computation and memory. Specifically, we aim to fuse core operators for faster execution and implement a targeted activation management strategy. This memory focus is essential: each additional stream introduces a proportional copy of intermediate activations, making efficient memory handling critical for scaling.

\subsubsection{Kernel Fusion}
As demonstrated by the flowchart \Cref{fig:mgr_flow} and pseudo code in \Cref{code:pytorch}, the output of the lerp operation is exactly the input to the attnpool at the same layer, these two operations can be completely fused into a single custom kernel.
The primary overhead bottleneck evidently stems from the memory accesses to the 
$n$-fold stream. Here, we provide a brief analysis. Assuming a fully fused operator for these two operations, the memory traffic for the 
$n$-fold stream minimally involves the following stages:
\begin{enumerate}
\item \textbf{1st RMSNorm}: Requires one read of size $nC$ to getting the input streams state. This step can be directly fused with the pre-computation of gating score.
\item \textbf{Lerping}: Involves one read and one write of size $nC$, which can be executed concurrently with the \texttt{new\_streams} rmsnorm calculation and the pre-computation of attnpool score.
\item \textbf{Attention Pooling}: Requires one read of size $nC$ for averaging \texttt{new\_h} from \texttt{new\_streams}.
\end{enumerate}
Roughly, the total I/O traffic amounts to $3nC$ reads and $1nC$ writes. This estimation is grounded in our actual implementation of the fused operators, serving as a validated practical lower bound. Although we have not deeply investigated more aggressive fusion patterns to further minimize the I/O, empirical results confirm that even at this current level of implementation, our approach already yields a clear and sufficient advantage over alternative architectures.

\subsubsection{Recomputing}
The $n$-stream residual design introduces substantial memory overhead during training. To mitigate this, we adopt a selective activation recomputation strategy akin to mHC, discarding the intermediate activations of lightweight operations while preserving the computed states of the heavy layer function $\mathcal{F}$.
mHC then employs an ingenious block-wise checkpointing strategy, significantly reducing the memory footprint by recomputing the stream states of individual layers within each block.

\begin{figure}[htbp]
  \centering
  \begin{subfigure}[b]{0.45\textwidth}
    \label{fig:cmgr_p1_hist}
    \vspace{0.5em}
    \includegraphics[width=\linewidth]{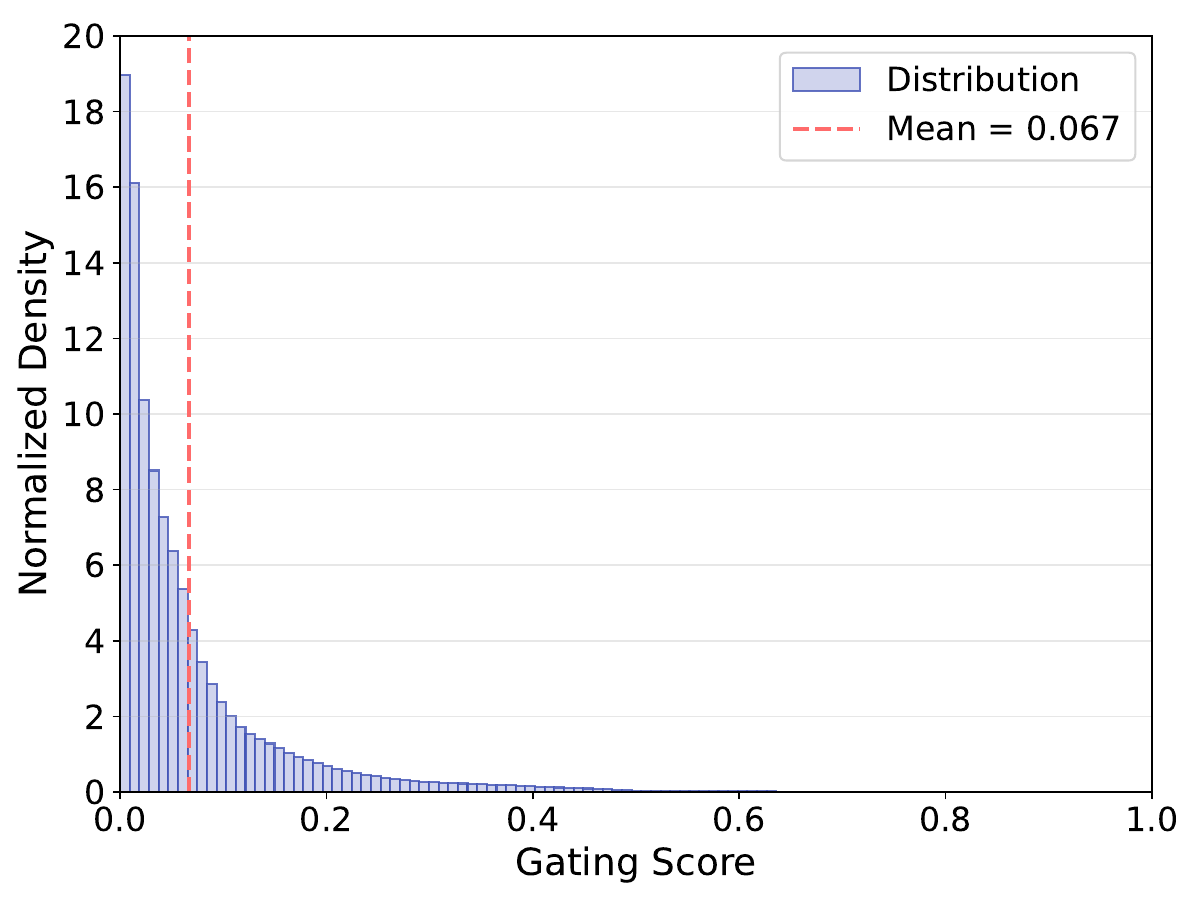}
  \end{subfigure}
  \hfill
  \begin{subfigure}[b]{0.45\textwidth}
    \label{fig:cmgr_p1_box}
    \vspace{0.5em}
    \includegraphics[width=\linewidth]{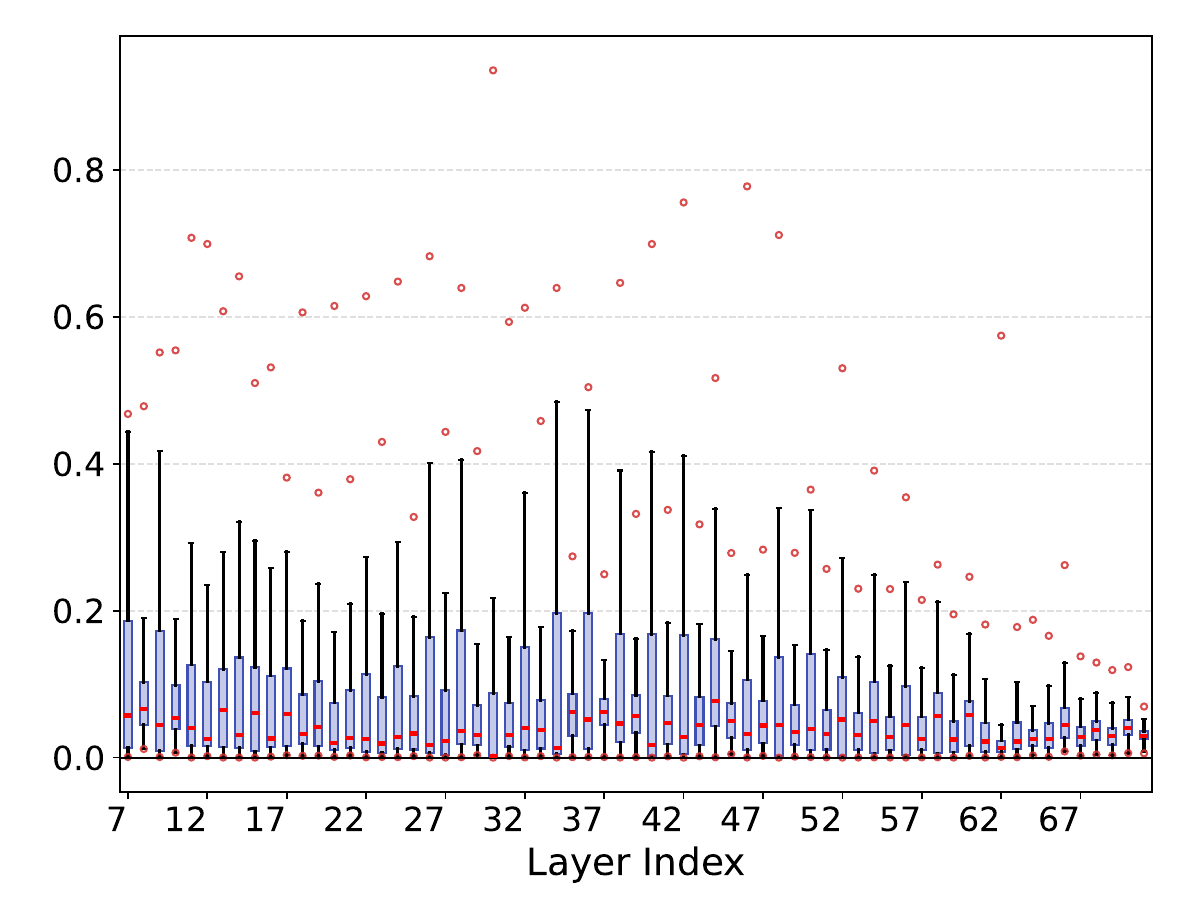}
  \end{subfigure}  
  \caption{Gating score statistics of the competitive MGR ($n=8$). Distributions and means of the gate values across all layers (Left), and layerwise box plots of the gate values (Right), the red hollow circles above the box represent the maximum values (outliers) in the corresponding layers.}
  \label{fig:cmgr_p1}
\end{figure}

However, we noticed that our calculation of next streams $\bm{s}'$ is achieved through a simple lerping processs (\Cref{eq:lerp}), under which the input streams $\bm{s}$ can also be easily reverse-solved from the output streams, as long as the gaiting socre $\beta$ is retained:
\begin{equation}
  \bm{s}_i =  (\bm{s}_i' - \beta_i \odot  \mathcal{F}_l\left(\bm{h}_l\right)) \odot \frac{1} {1-\beta_i}
\label{eq:re_lerp}
\end{equation}

where $\mathcal{F}_l\left(\bm{h}_l\right)$ is the layer's output which is already retained for backpropagation.
During backpropagation, gradients flow from the final layer toward the first layer. By employing some scheme in which the inversed input streams of the current layer is passed to the preceding layer during the backward pass, we can always sequentially recompute the input 
stream of every layer across the entire computation graph while storing only the last stream state for MGR.
The inverse solution of this equation can be reliably obtained only when the interpolation coefficient ($\beta$) is sufficiently small. Otherwise, the accuracy of the inverse solution becomes difficult to guarantee.  To verify this, we visualized the distribution of $\beta$ values across all actual layers and plotted the results in \Cref{fig:cmgr_p1} and \Cref{fig:imgr_p1}. As shown, the actual $\beta$ values are usually very small on average, with the majority falling below $0.1$. Only a few outliers reach the range of $0.8$--$0.9$.

Therefore, we adopt the following \textbf{Fallback Invertion} strategy:  
During the forward pass, based on the gate values $\beta$, we keep only the input streams  $\tilde{\bm{s}}$ that correspond to the top-$p$ largest values.  
In the backward pass, we first inversely solve for the input $\bm{s}$ from the returned output, and then fallback to the saved $\tilde{\bm{s}}$ at those recorded indices.  
This achieves a relatively accurate acquisition of the input stream state while saving GPU memory.
On the deepest models we tested, we compared this Fallback Inversion algorithm ($p=0.01$) with the standard backpropagation algorithm that requires storing input streams on the computation graph. The training curves and validation set performance were virtually identical.

\section{Conclusion and Discussion}

In this work, we propose Multi-Gate Residuals (MGR), a novel residual structure for neural networks that maintains feature representations through multiple streams. These streams are updated via a lightweight gating mechanism, while an attention pooling (AttnPool) module is employed to aggregate information from the streams, producing compact features for subsequent layers. Overall, our design innovatively combines core components from several prior works \citep{srivastava_highway_2015, team_attention_2026, zhu_hyper-connections_2025}.
Under this architecture, we introduce two variants and evaluate them across three model scales, defined by their layer counts: 12, 24, and 36. In all cases, both variants consistently and substantially outperform competing architectures of similar sizes.
Furthermore, based on the observed feature characteristics of the model, we designed tailored engineering strategies. These optimizations ensure that our approach introduces no communication bottlenecks in parallel computing while achieving excellent computational and memory efficiency. As a result, the proposed method is highly practical for large-scale deployment, incurring only negligible training overhead and minimal inference overhead.

We present a distinct philosophical shift from contemporary architectures that focus on designing complex, hand-crafted projection operators for stream state updates. While those methods attempt to engineer a richer semantic space through intricate spatial transformations, we argue that such approaches risk overriding the high-level representations painstakingly built by the core non-linear layers.
MGR takes a fundamentally different path that closely aligns with the principle of Slot Attention \citep{locatello_object-centric_2020}. By treating parallel stream states as co-equal memories, we eliminate the need for complex, top-down transformations. Instead, each stream independently updates its semantics using a straightforward scoring and gating mechanism. This approach ensures that the primary layers retain their role as the exclusive feature extractors, while the streams function efficiently as parallel, undistorted memory banks.

MGR achieves superior performance, notably surpassing AttnRes. This improvement is particularly significant given that Full AttnRes employs a deepwise attention mechanism that maintains access to outputs from arbitrary historical layers--—a design choice that inevitably incurs substantial communication overhead. We speculate this full-history attention may impair its capacity to focus on truly salient features.
%In fact, their analysis of the depth-wise attention weights already shows that attention in the depth direction tends to focus primarily on more recent history, suggesting that full-depth attention may not be strictly necessary. 
We leave a more rigorous validation of this hypothesis and confirming whether our method maintains its advantages at larger scales to future work.

\bibliographystyle{unsrtnat}
\bibliography{references}  %%% Uncomment this line and comment out the ``thebibliography'' section below to use the external .bib file (using bibtex) .

\clearpage
\appendix
\section{PyTorch-style pseudo code for Multi-Gate Residuals}

\begin{lstlisting}[caption={Pseudo code for MGR. Argument  \texttt{gate\_variant} is used to switch between different gating mechanisms.}, label={code:pytorch}, float=htbp]
class MultiGateResidual(nn.Module):
    def __init__(
        self, d_model, n_stream, gate_variant="competitive", init_bias=0, eps=1e-6
    ):
        super().__init__()
        assert gate_variant in ["competitive", "independent"]
        self.d_model = d_model
        self.n_stream = n_stream
        self.eps = eps
        self.scale = d_model**-0.5
        self.w1 = nn.Parameter(torch.zeros(d_model))
        self.w2 = nn.Parameter(torch.zeros(d_model))

        if gate_variant == "competitive":
            self.b1 = nn.Parameter(torch.zeros(n_stream + 1))
            self.b1.data[0] = init_bias
        else:
            self.b1 = nn.Parameter(torch.zeros(n_stream))
            nn.init.constant_(self.b1, init_bias)
        self.gate_variant = gate_variant

    def forward(self, layer_output, streams):
        # layer_output: [B, T, D]
        # streams: [B, T, N, D]
        normed_s = F.rms_norm(streams, (self.d_model,), eps=self.eps)
        score_s = (normed_s * self.w1).sum(-1) * self.scale
        if self.gate_variant == "competitive":
            logit = F.pad(score_s, (1, 0), value=0) + self.b1
            beta = logit.softmax(dim=-1)[..., 1:]
        else:
            logit = score_s + self.b1
            beta = logit.sigmoid()
        new_streams = streams + beta.unsqueeze(-1) * (
            layer_output.unsqueeze(-2) - streams
        )

        normed_ns = F.rms_norm(new_streams, (self.d_model,), eps=self.eps)
        score_ns = (normed_ns * self.w2).sum(-1) * self.scale
        alpha = score_ns.softmax(dim=-1)
        new_h = torch.einsum("btn,btnd->btd", alpha, new_streams)

        return new_h, new_streams
\end{lstlisting}
\FloatBarrier

\section{Hyperparameters}
\label{sec:hyperp}
Our implementation is based on nanoGPT. 
Our primary structural deviation is the adoption of $\text{ReLU}^2$ \citep{so2022primersearchingefficienttransformers} as the FFN activation, which preliminary tests show yields consistent gains over the baseline.
All models are trained from scratch using a dual-parameter-group setup. We apply the Muon optimizer to all 2D weight matrices and AdamW to biases, normalization layers, and embeddings. A global cosine learning rate schedule with linear warmup is used throughout.
For the three model scales (S, M, and L), their scale-specific hyperparameters listed in \Cref{tab:specific-hyper-param}. 

\begin{table}[htbp]
\caption{Architectures and learning rates. All models are trained with a batch size of $0.5M$ tokens, for a total of 200K iterations.}
% \vspace{0.6em}
\centering
\renewcommand{\arraystretch}{1.2}
\begin{tabular}{ccccccccc}
\toprule
Scale Size
& $n_{\text{layers}}$
& $d_{\text{model}}$
& $n_{\text{heads}}$
& $d_{\text{head}}$
& lr (AdamW) 
& lr (Muon) \\
\midrule
S (0.12B)
& 12
& 768
& 6
& 128
& $3.0 \times 10^{-3}$
& $1.0 \times 10^{-2}$ \\
\midrule
M (0.35B)
& 24
& 1024
& 8
& 128
& $2.0 \times 10^{-3}$
& $6.0 \times 10^{-3}$ \\
\midrule
L (0.77B)
& 36
& 1280
& 10
& 128
& $1.0 \times 10^{-3}$
& $4.0 \times 10^{-3}$ \\

\bottomrule
\end{tabular}
% \vspace{0.6em}
\label{tab:specific-hyper-param}
\end{table}

\FloatBarrier

The shared hyperparameters used across all experiments are summarized in \Cref{tab:shared-hyper-param}.

\begin{table}[htbp]
    \caption{Shared hyperparameters.}
    % \vspace{0.6em}
    \centering
    \begin{tabular}{l|c}
    \toprule
       Name  & Value \\ \midrule
       Batch Size  & 512K tokens \\ 
       Block Size (Sequence Length)  & 1024 \\
       Total Iterations & 20000 \\
       Warmup Iterations & 200 \\
       weight decay & 0.1 \\ 
       Learning Rate Schedule & Cosine Decay \\
       Adam $\beta$ & (0.9, 0.95) \\
       Muon $\mu$ & 0.95 \\
       Grad Clip & 1.0 \\ 
       Dropout & 0.0 \\ 
       Tie Word Embedding & True \\
       Activation Function  & $\text{ReLU}^2$ \\
       Position Embedding & RoPE ($\theta=10000$) \\
       \bottomrule
    \end{tabular}
    % \vspace{0.6em}
    \label{tab:shared-hyper-param}
\end{table}

\FloatBarrier

\needspace{0.65\textheight} 
\section{Additional Figures}
\subsection{Extended Results for Layerwise Augular Distance}
\label{sec:ext_layerwise_sim}
\begin{figure}[htbp]
  \centering
  \begin{subfigure}[b]{0.32\textwidth}
	\caption{}
    \label{fig:heatmap_cmgr1}
    \vspace{0.5em}
    \includegraphics[width=\linewidth]{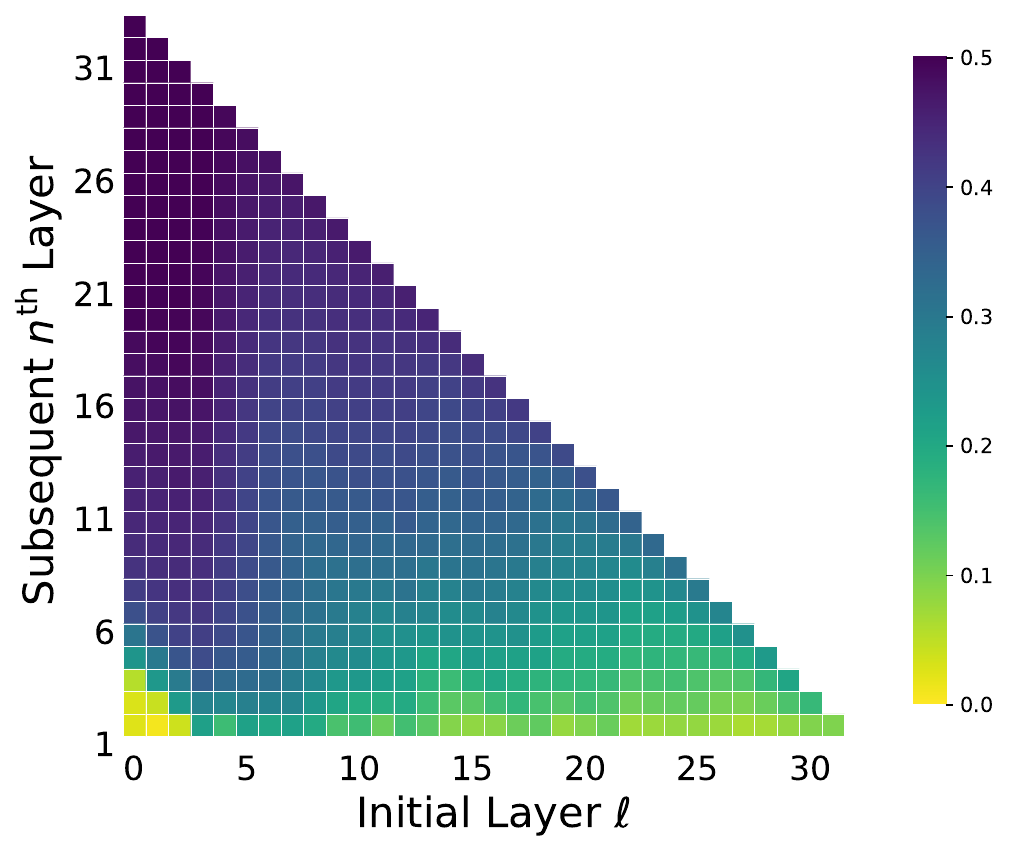}
  \end{subfigure}
  \hfill
  \begin{subfigure}[b]{0.32\textwidth}
	\caption{}
    \label{fig:heatmap_cmgr4}
    \vspace{0.5em}
    \includegraphics[width=\linewidth]{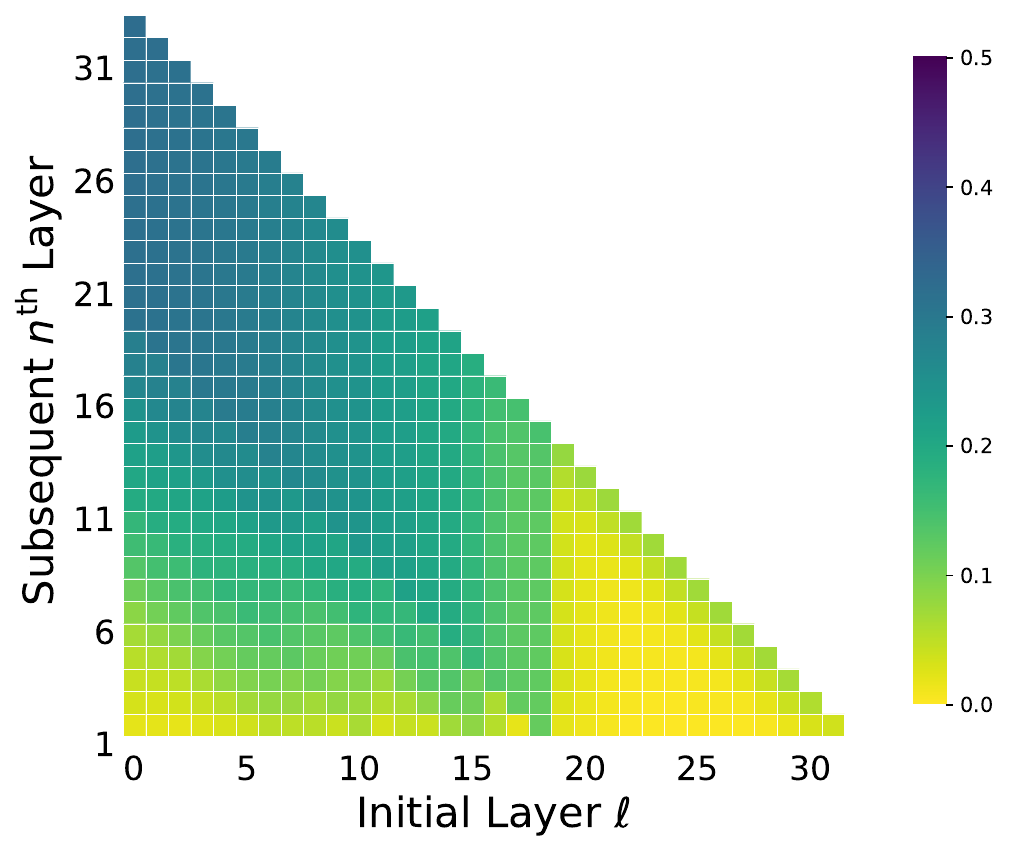}
  \end{subfigure}
  \hfill
  \begin{subfigure}[b]{0.32\textwidth}
	\caption{}
    \label{fig:heatmap_cmgr5}
    \vspace{0.5em}
    \includegraphics[width=\linewidth]{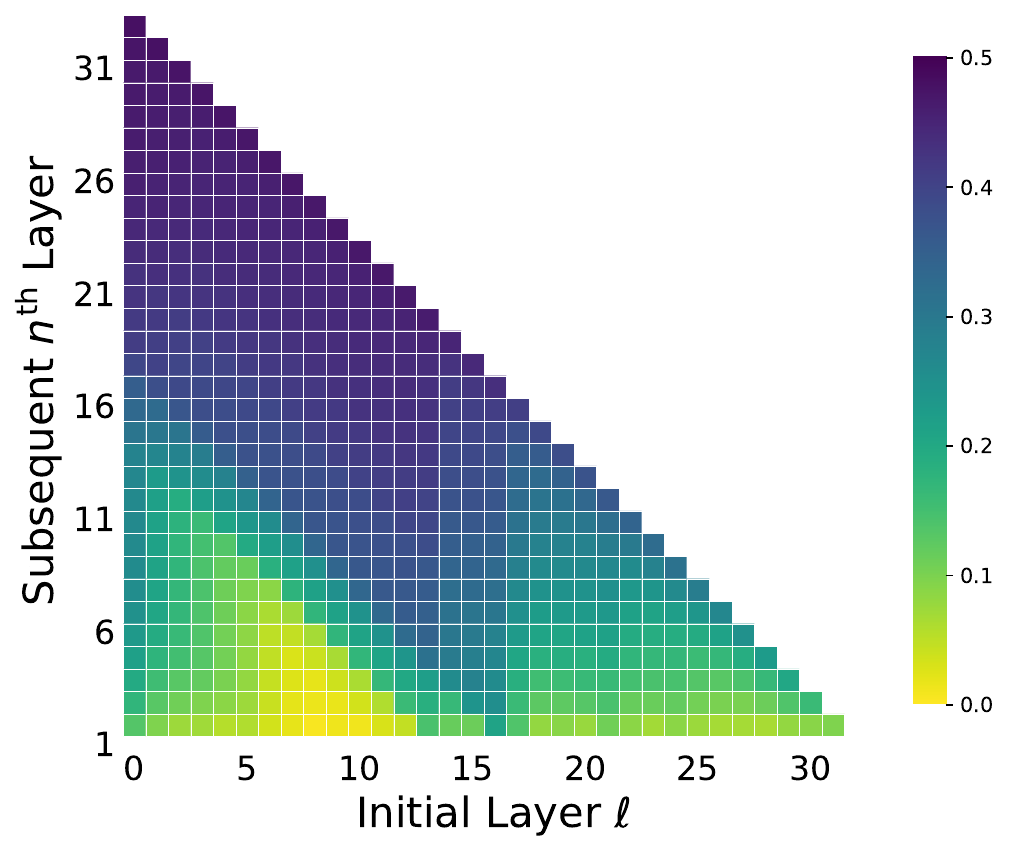}
  \end{subfigure}
  \hfill
  \begin{subfigure}[b]{0.32\textwidth}
	\caption{}
    \label{fig:heatmap_cmgr6}
    \vspace{0.5em}
    \includegraphics[width=\linewidth]{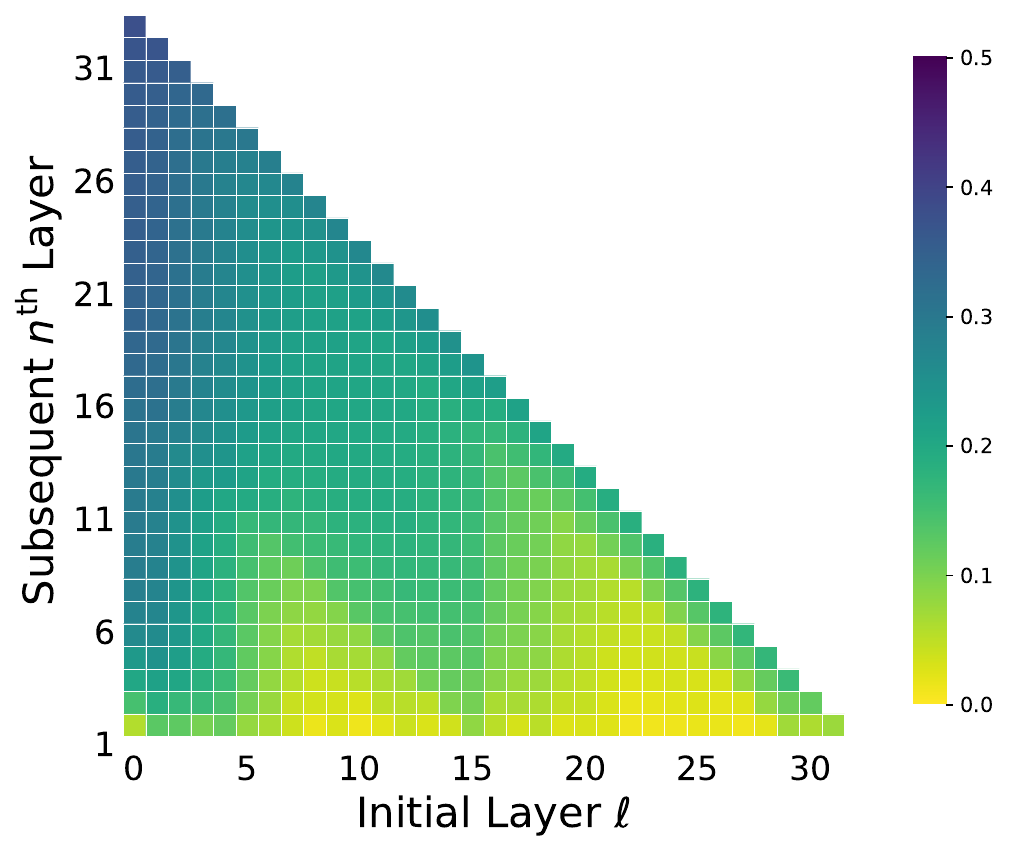}
  \end{subfigure}
  \qquad
  % \hfill
  \begin{subfigure}[b]{0.32\textwidth}
	\caption{}
    \label{fig:heatmap_cmgr7}
    \vspace{0.5em}
    \includegraphics[width=\linewidth]{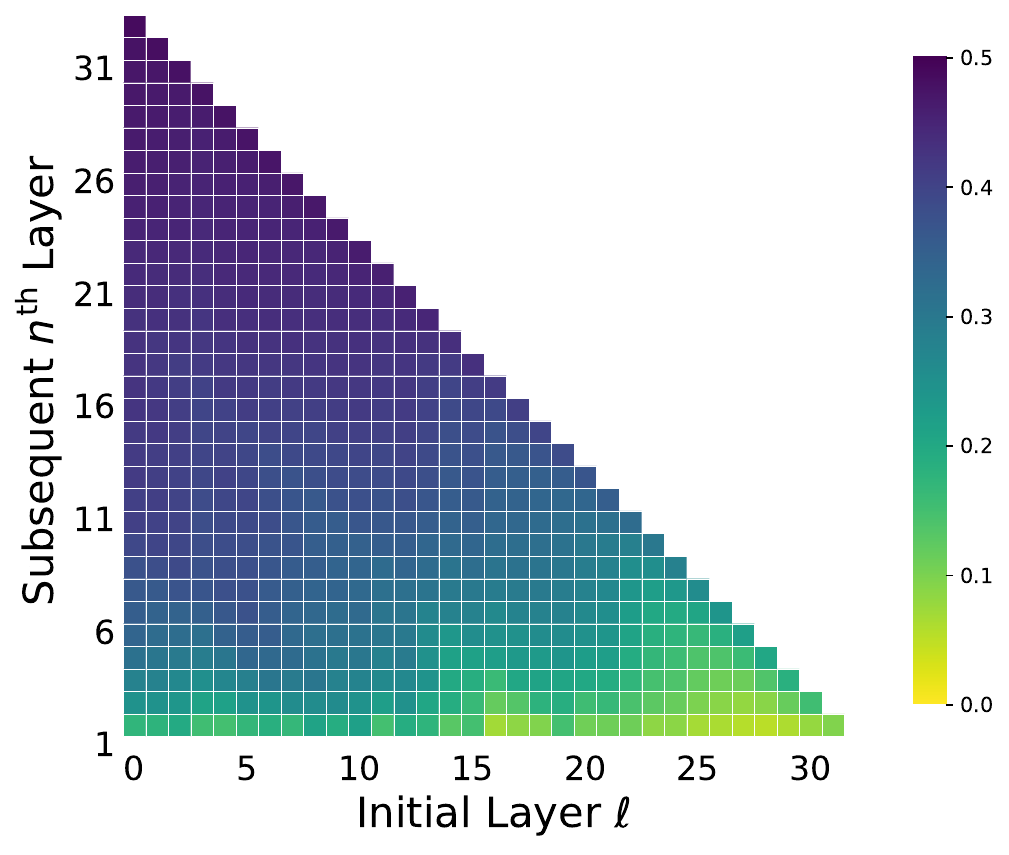}
  \end{subfigure}
  \caption{Angular distance from the initial block $\ell$ (x-axis) and its subsequent $n^{\mathrm{th}}$ block (y-axis). (a) augular distance heatmap from Pre-Norm Architecture, (b-e) augular distance heatmap of the Competitive MGR ($n=8$), showing feature similarity derived from the streamwise indexed
   as 1, 4, 5, 6, 7, in that order.
  }
  \label{fig:heatmap_more_cmgr}
\end{figure}

\begin{figure}[htbp]
  \centering
  \begin{subfigure}[b]{0.32\textwidth}
	\caption{}
    \label{fig:heatmap_imgr0}
    \vspace{0.5em}
    \includegraphics[width=\linewidth]{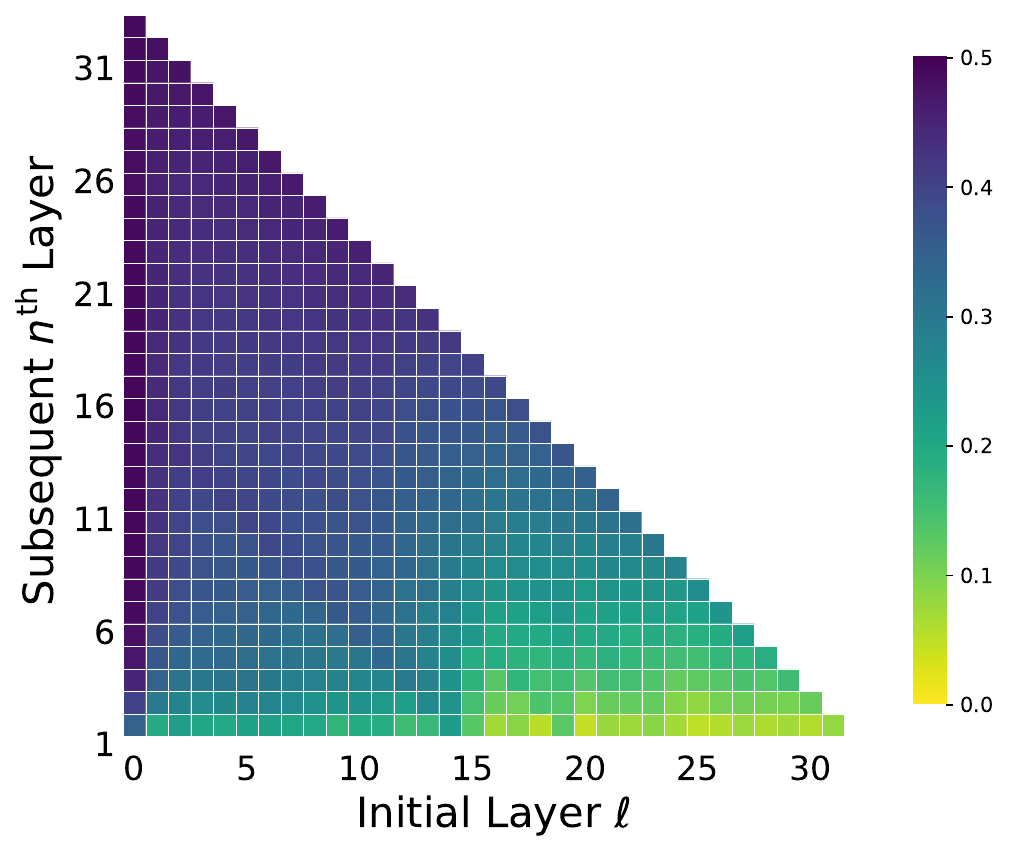}
  \end{subfigure}
  \hfill
  \begin{subfigure}[b]{0.32\textwidth}
	\caption{}
    \label{fig:heatmap_imgr1}
    \vspace{0.5em}
    \includegraphics[width=\linewidth]{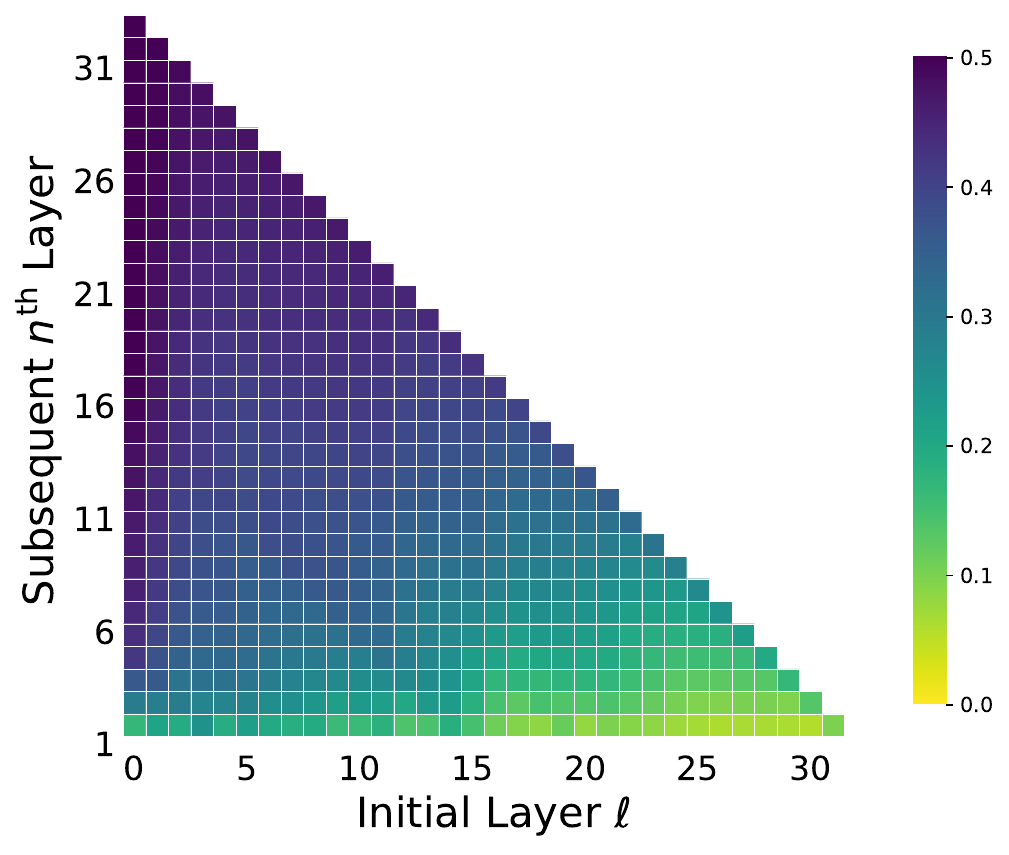}
  \end{subfigure}
  \hfill
  \begin{subfigure}[b]{0.32\textwidth}
	\caption{}
    \label{fig:heatmap_imgr2}
    \vspace{0.5em}
    \includegraphics[width=\linewidth]{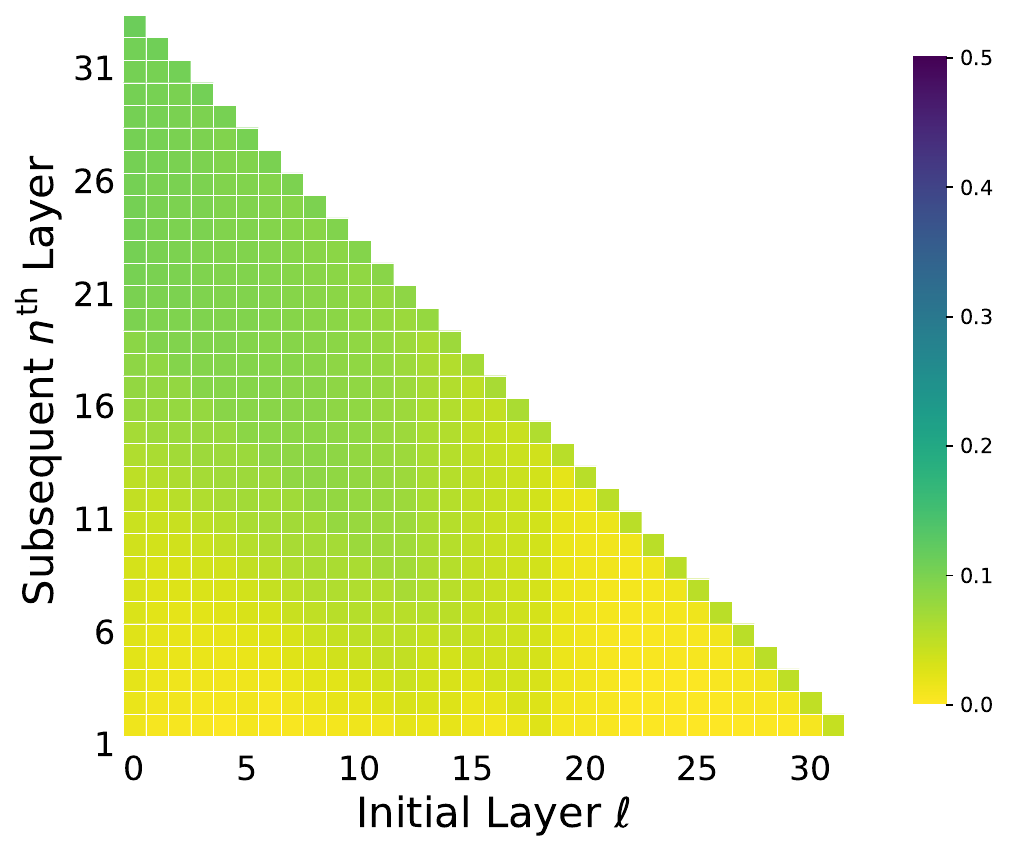}
  \end{subfigure}
  \hfill
  \begin{subfigure}[b]{0.32\textwidth}
	\caption{}
    \label{fig:heatmap_imgr3}
    \vspace{0.5em}
    \includegraphics[width=\linewidth]{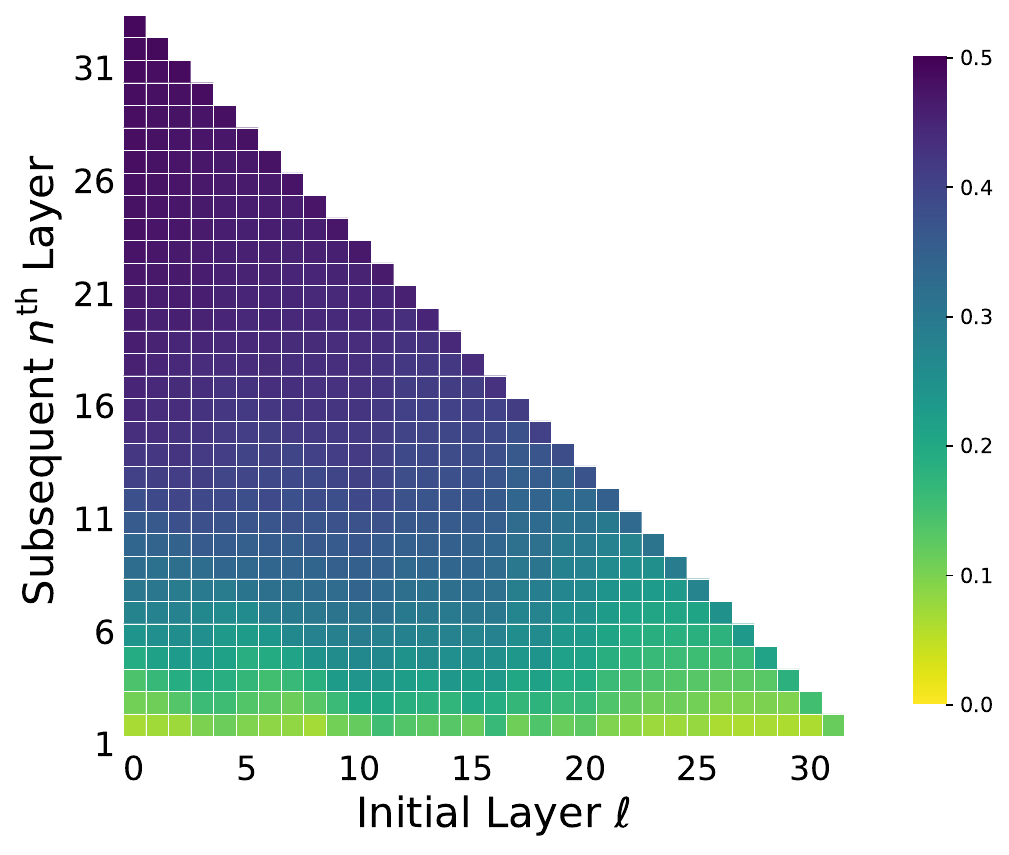}
  \end{subfigure}
  \hfill
  \begin{subfigure}[b]{0.32\textwidth}
	\caption{}
    \label{fig:heatmap_imgr4}
    \vspace{0.5em}
    \includegraphics[width=\linewidth]{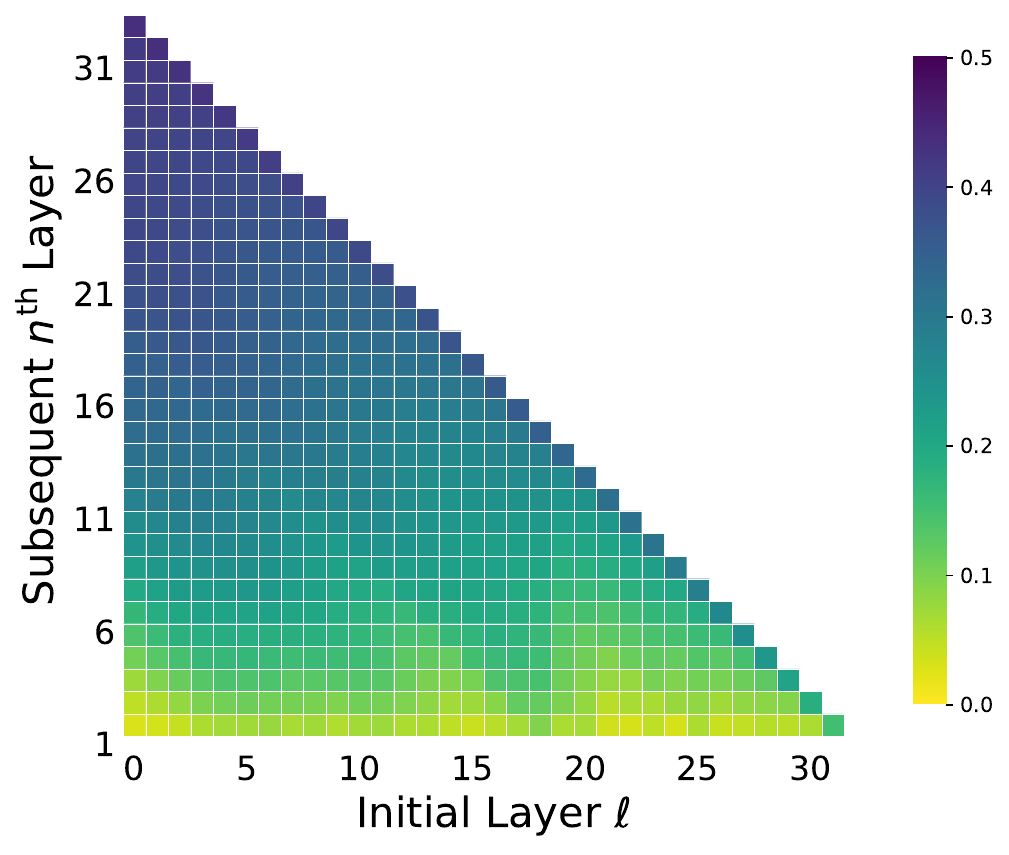}
  \end{subfigure}
  \hfill
  \begin{subfigure}[b]{0.32\textwidth}
	\caption{}
    \label{fig:heatmap_imgr5}
    \vspace{0.5em}
    \includegraphics[width=\linewidth]{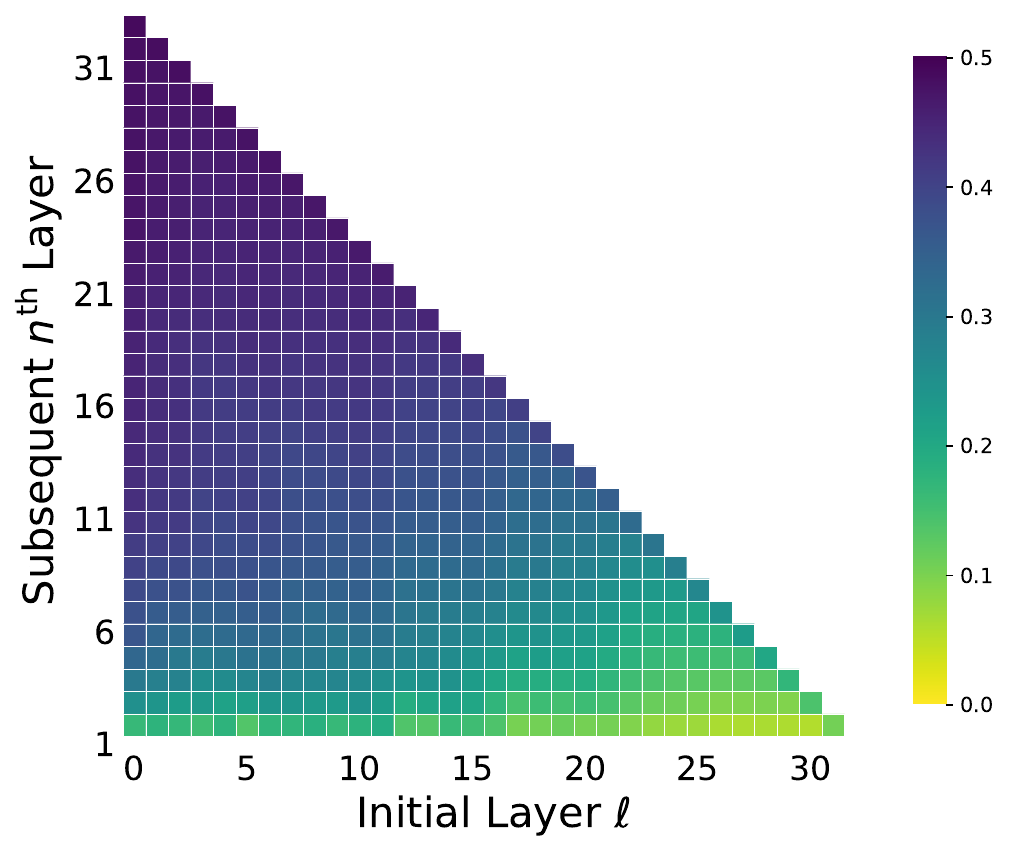}
  \end{subfigure}
  \hfill
  \begin{subfigure}[b]{0.32\textwidth}
	\caption{}
    \label{fig:heatmap_imgr6}
    \vspace{0.5em}
    \includegraphics[width=\linewidth]{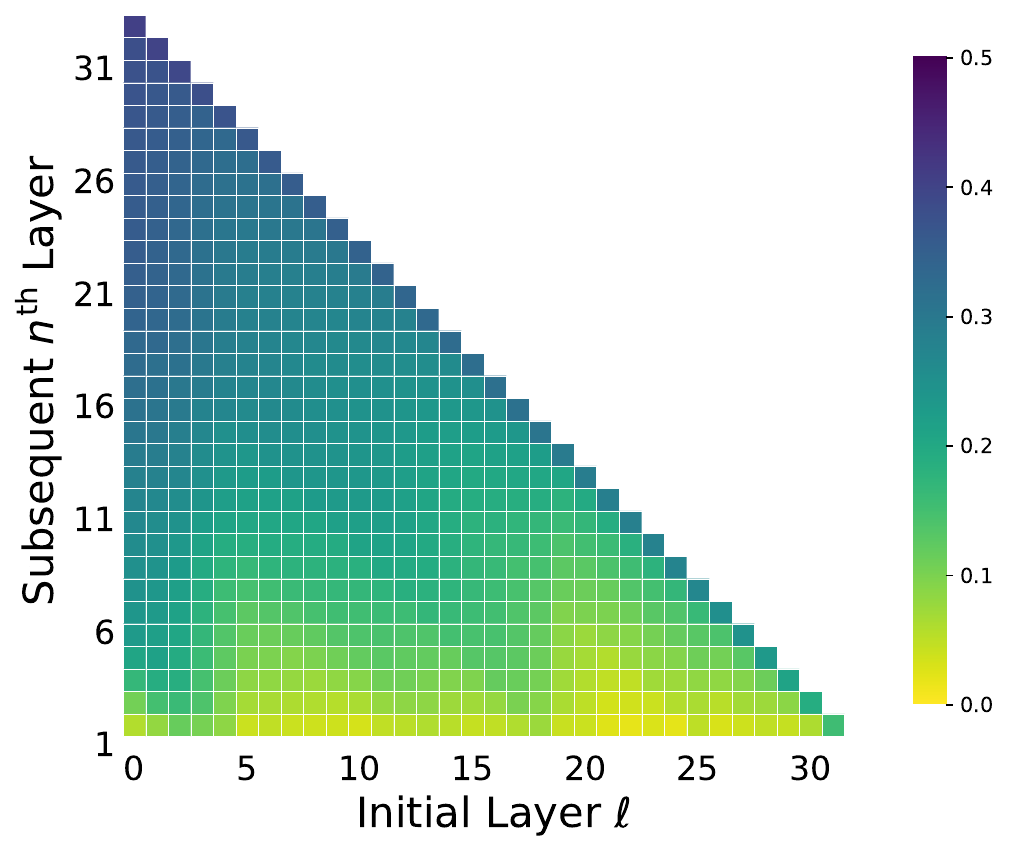}
  \end{subfigure}
  % \hfill
  \qquad
  \begin{subfigure}[b]{0.32\textwidth}
	\caption{}
    \label{fig:heatmap_imgr7}
    \vspace{0.5em}
    \includegraphics[width=\linewidth]{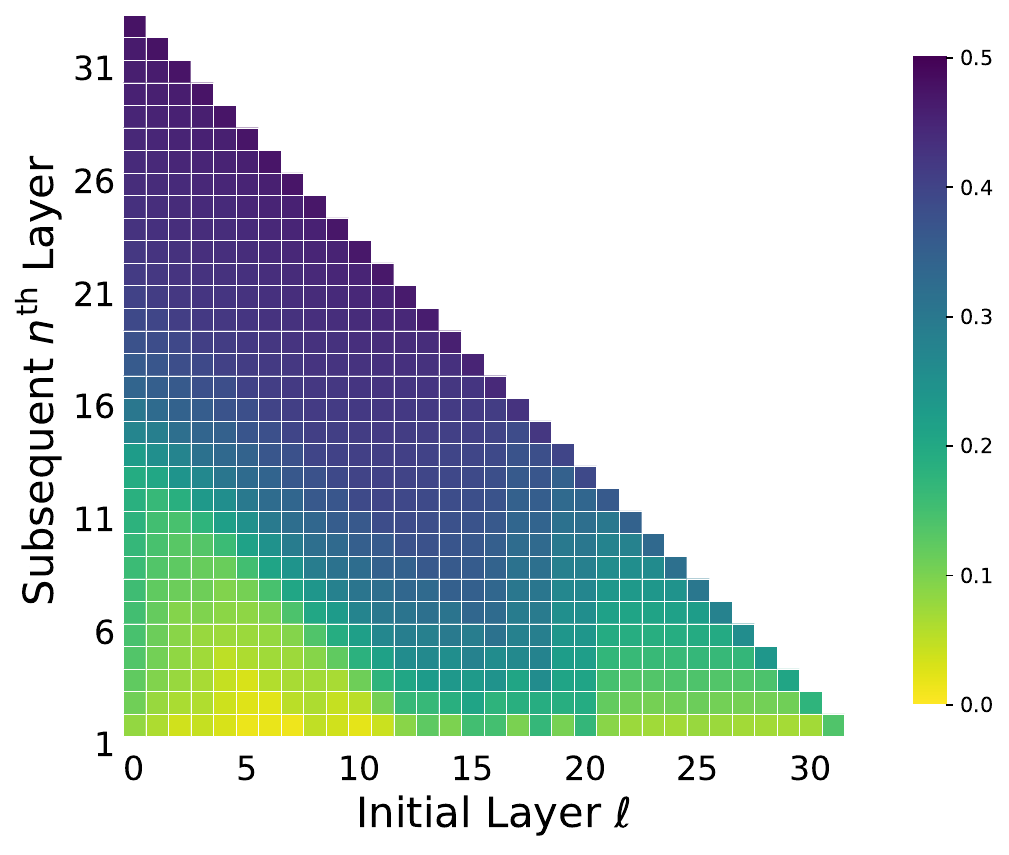}
  \end{subfigure}
  \caption{Angular distance from the initial block $\ell$ (x-axis) and its subsequent $n^{\mathrm{th}}$ block (y-axis). (a)-(h) shows feature similarity derived from each of the  8 streams of the independent MGR model.}
  \label{fig:heatmap_imgr}
\end{figure}

\FloatBarrier

\needspace{0.25\textheight} 
\subsection{Extended Results for Massive Activations}
\begin{figure}[htbp]
  \centering
  \begin{subfigure}[b]{0.32\textwidth}
	\caption{}
    \label{fig:ma_attn_preln}
    \vspace{0.5em}
    \includegraphics[width=\linewidth]{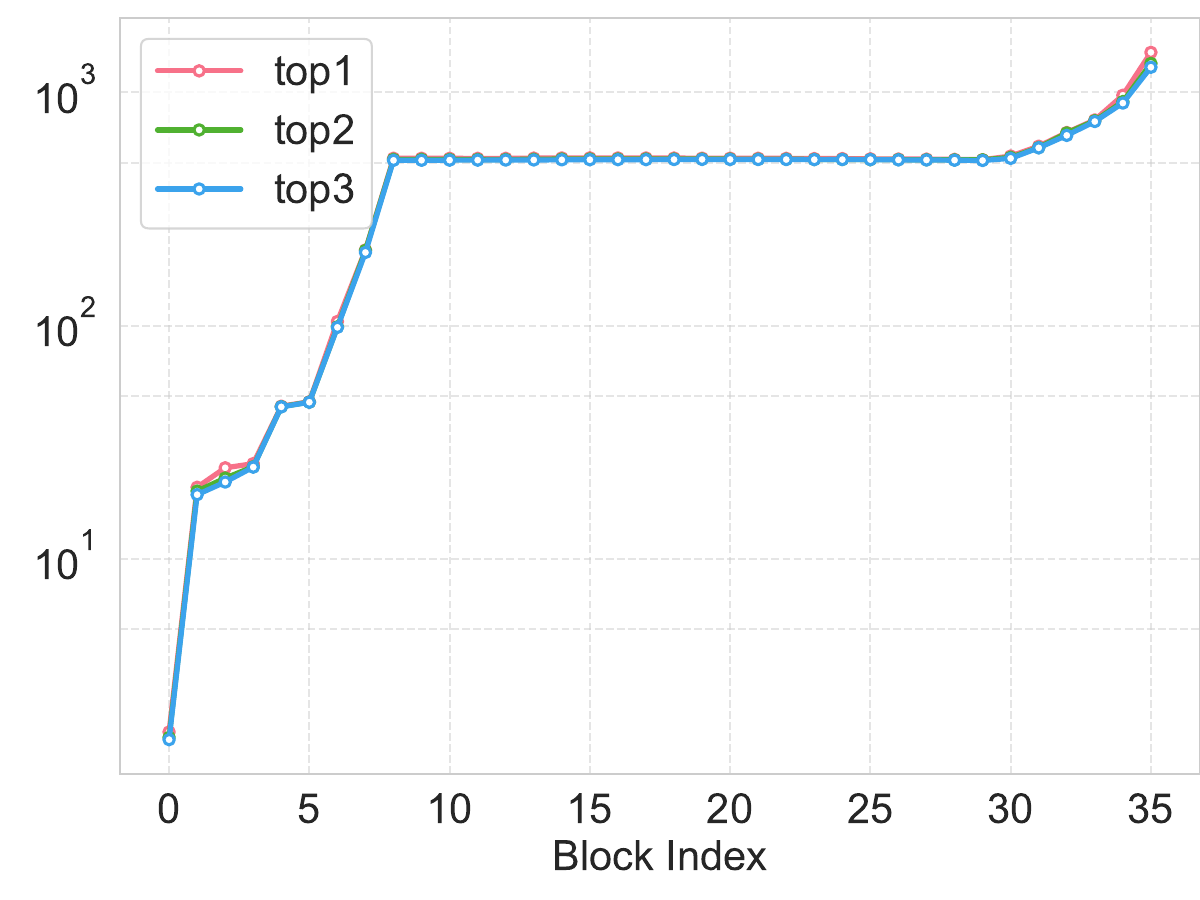}
  \end{subfigure}
  \hfill
  \begin{subfigure}[b]{0.32\textwidth}
	\caption{}
    \label{fig:ma_attn_fullattnres}
    \vspace{0.5em}
    \includegraphics[width=\linewidth]{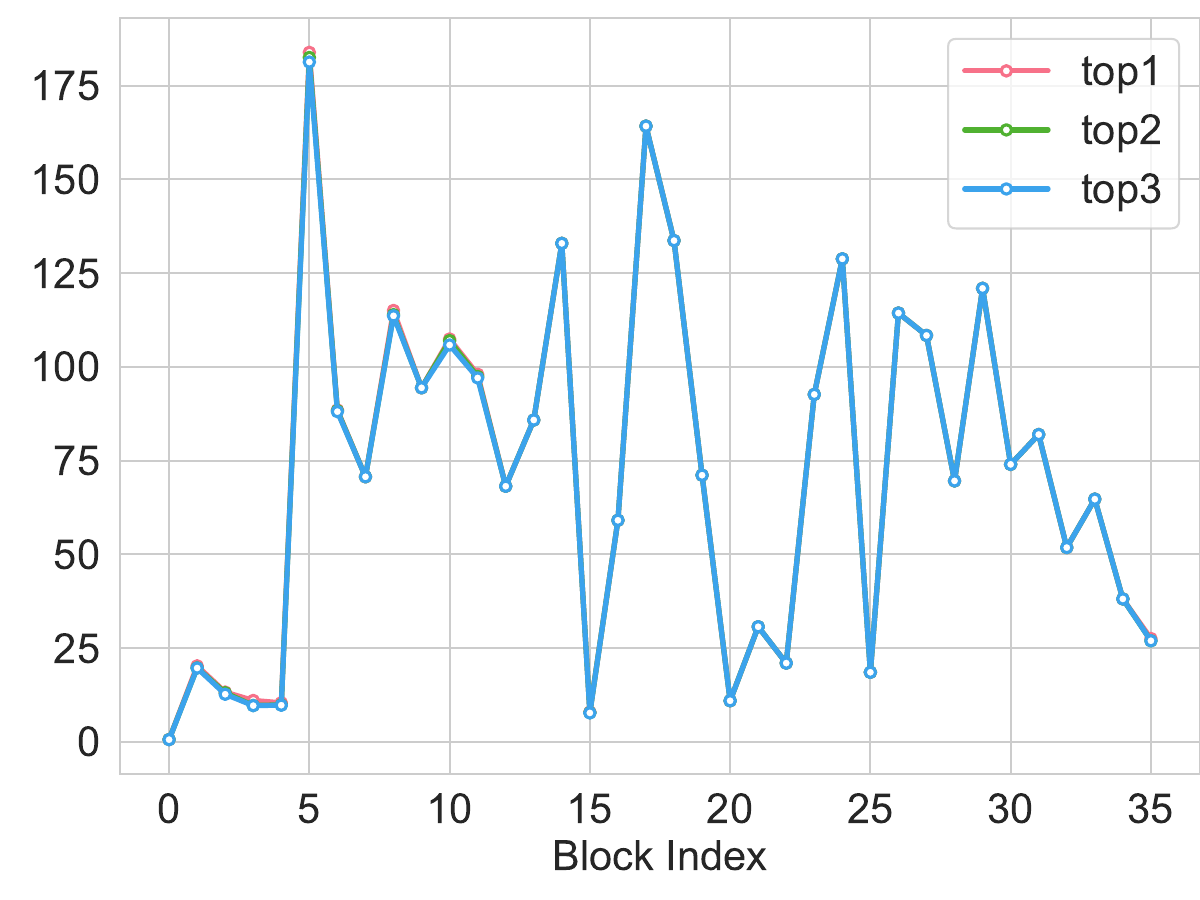}
  \end{subfigure}
  \hfill
  \begin{subfigure}[b]{0.32\textwidth}
	\caption{}
    \label{fig:ma_attn_cmgr}
    \vspace{0.5em}
    \includegraphics[width=\linewidth]{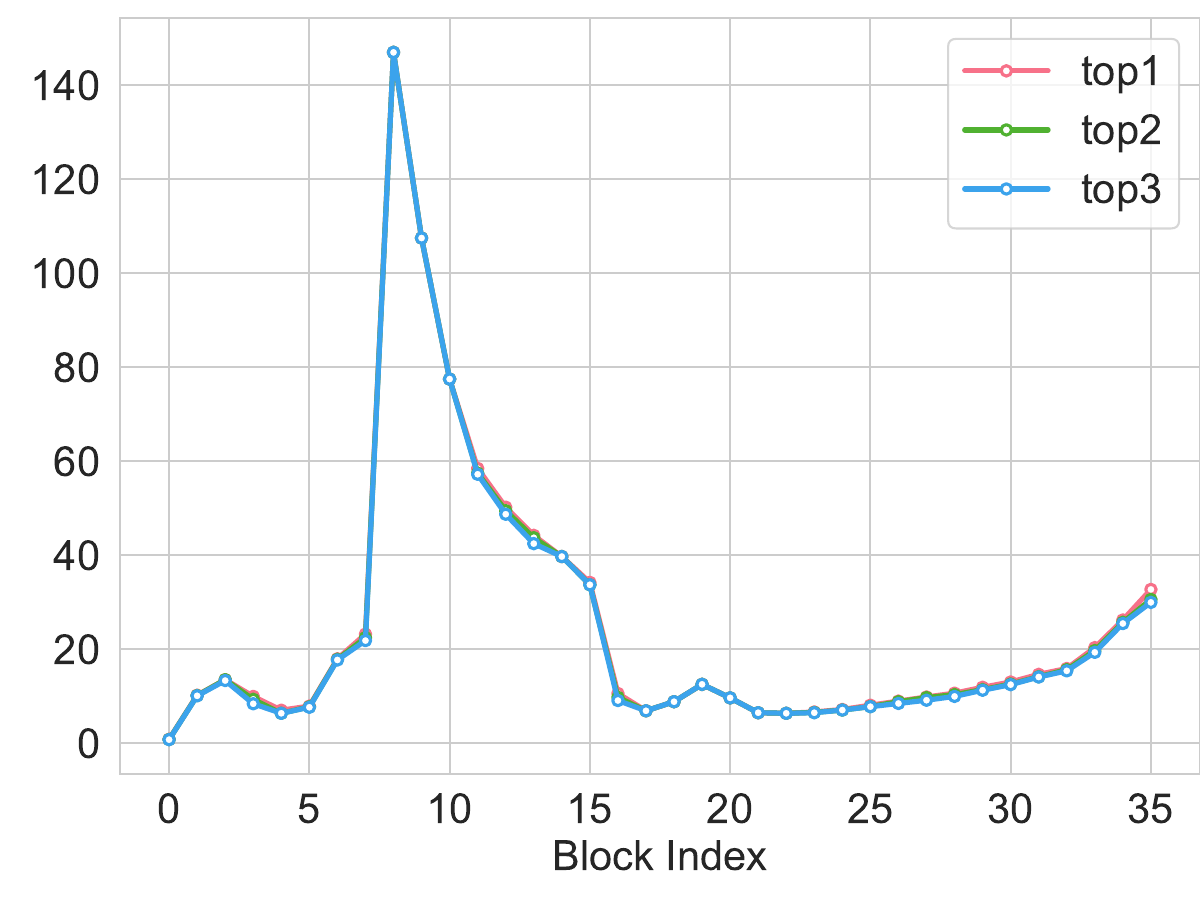}
  \end{subfigure}
  \caption{Maximum absolute values of the attention outputs  from (a) Pre-Norm model,  (b) Full AttnRes model and (c) Competitive MGR model.}
  \label{fig:ma_attn}
\end{figure}

\begin{figure}[htbp]
  \centering
  \begin{subfigure}[b]{0.32\textwidth}
	\caption{}
    \label{fig:ma_attn_imgr}
    \vspace{0.5em}
    \includegraphics[width=\linewidth]{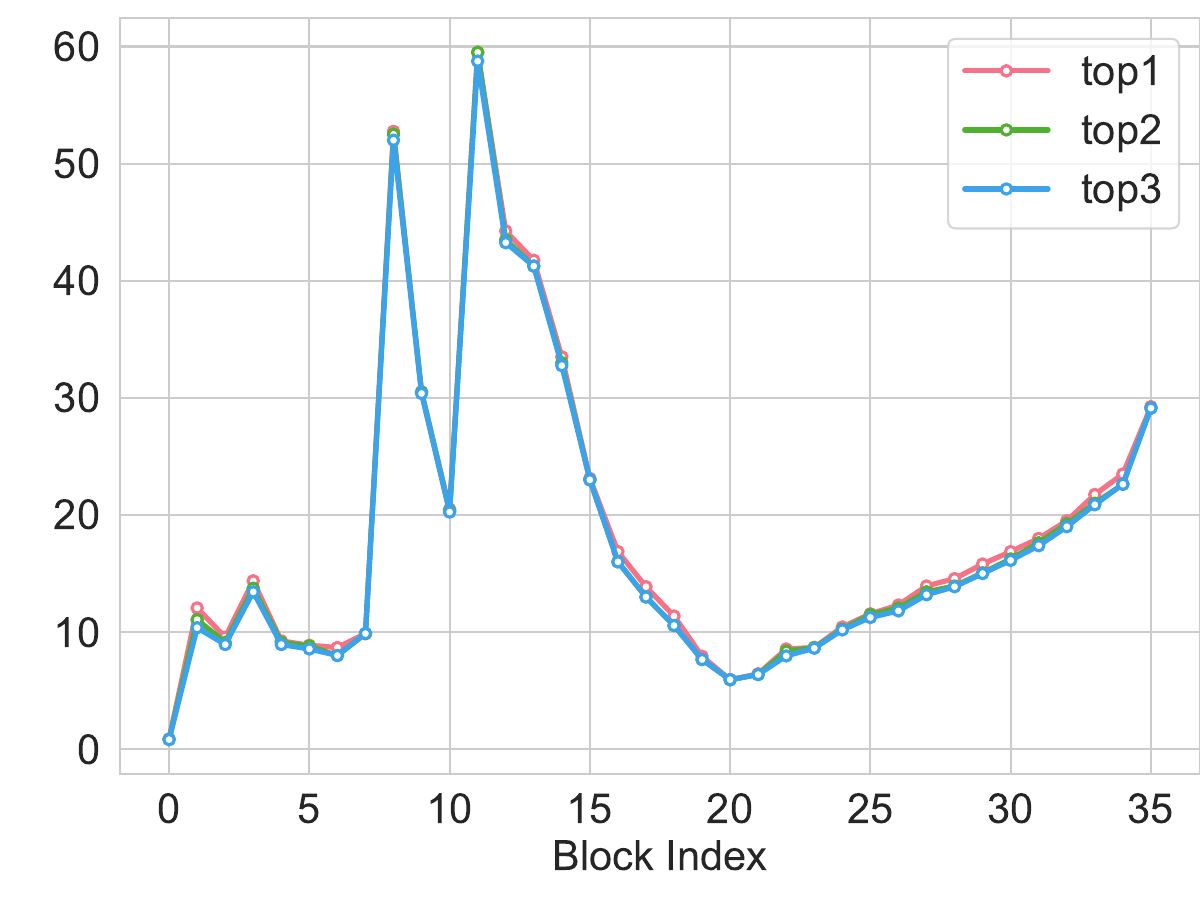}
  \end{subfigure}
  % \hfill
  \qquad
  \begin{subfigure}[b]{0.32\textwidth}
	\caption{}
    \label{fig:ma_attn_imgr}
    \vspace{0.5em}
    \includegraphics[width=\linewidth]{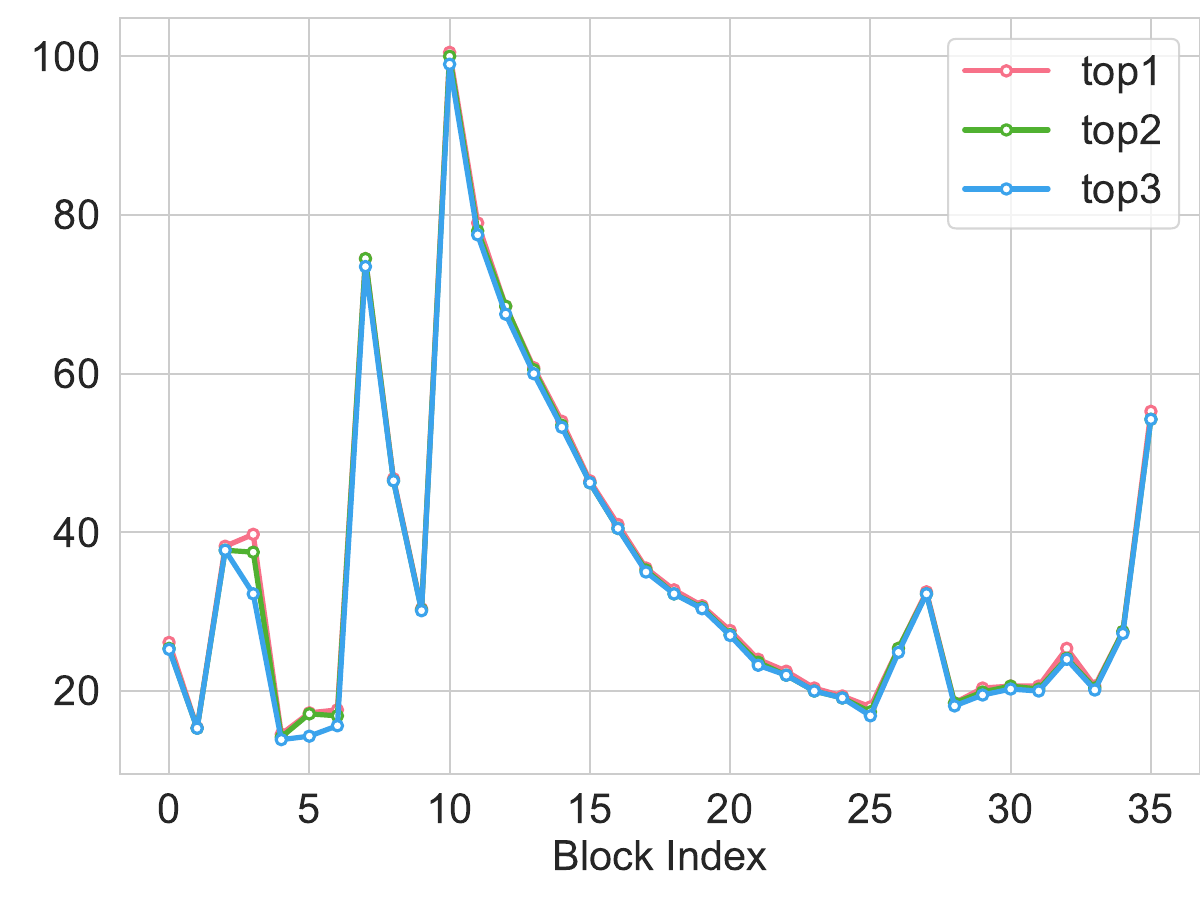}
  \end{subfigure}
  \caption{Maximum absolute values of the independent MGR model, with attention outputs shown in (a) and feedforward outputs shown in (b).}
  \label{fig:ma_imgr}
\end{figure}

\begin{figure}[htbp]
  \centering
  \begin{subfigure}[b]{0.45\textwidth}
	%\caption{}
    %\label{fig:ma_stream_attn_umgr}
    \vspace{0.5em}
    \includegraphics[width=\linewidth]{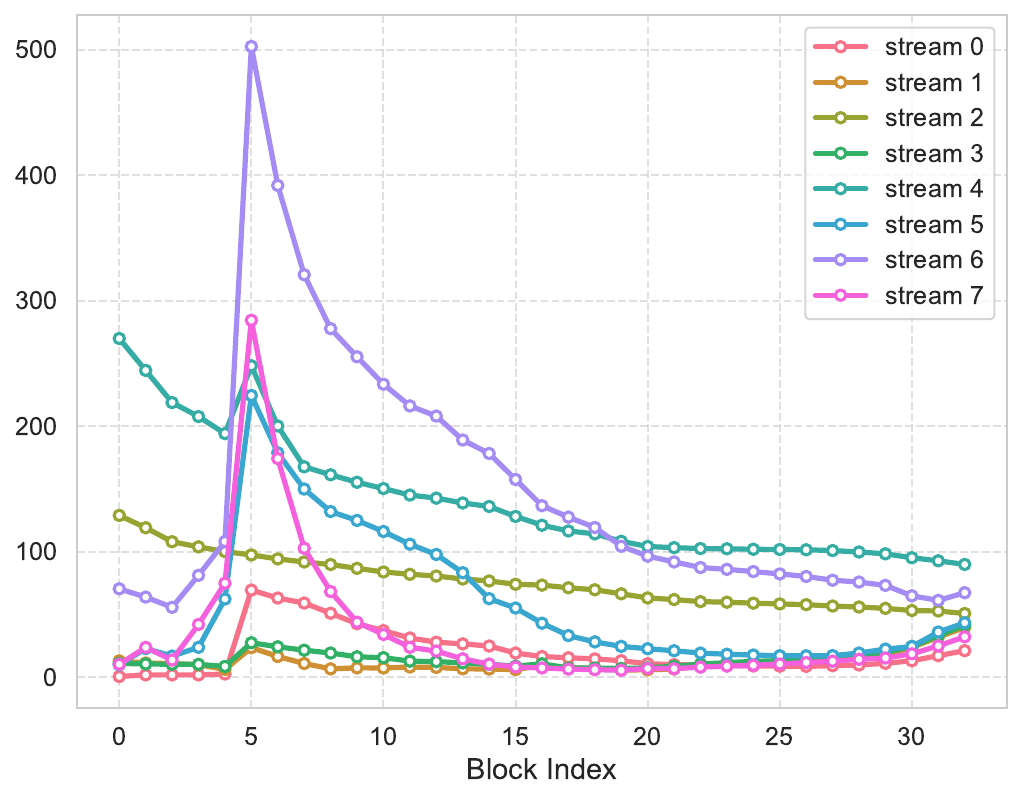}
  \end{subfigure}
  \hfill
  \begin{subfigure}[b]{0.45\textwidth}
	%\caption{}
    %\label{fig:ma_stream_attn_imgr}
    \vspace{0.5em}
    \includegraphics[width=\linewidth]{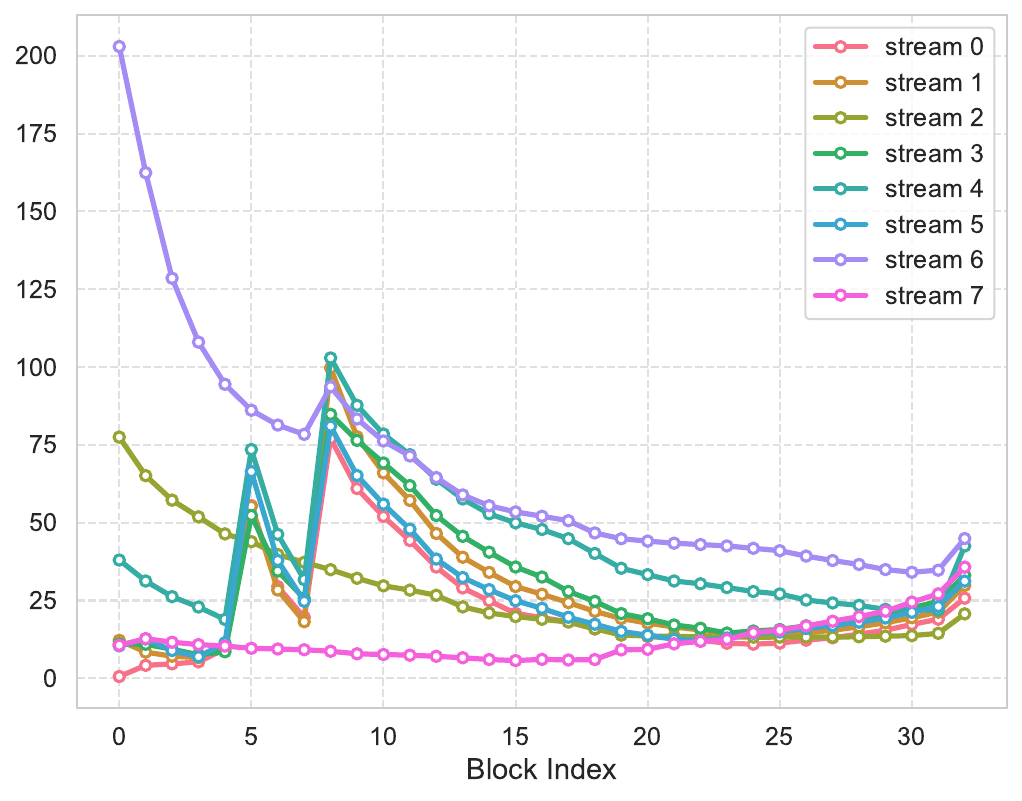}
  \end{subfigure}
  \caption{Maximum absolute value across all output streams for each attention layer. (Left): competitive MGR, (Right): independent MGR.}
  \label{fig:ma_stream_attn_mgr}
\end{figure}

\FloatBarrier

\needspace{0.3\textheight} 
\subsection{Extended Results for Gating Score Statistics}
\begin{figure}[htbp]
  \centering
  \begin{subfigure}[b]{0.45\textwidth}
    \label{fig:imgr_p1_hist}
    \vspace{0.5em}
    \includegraphics[width=\linewidth]{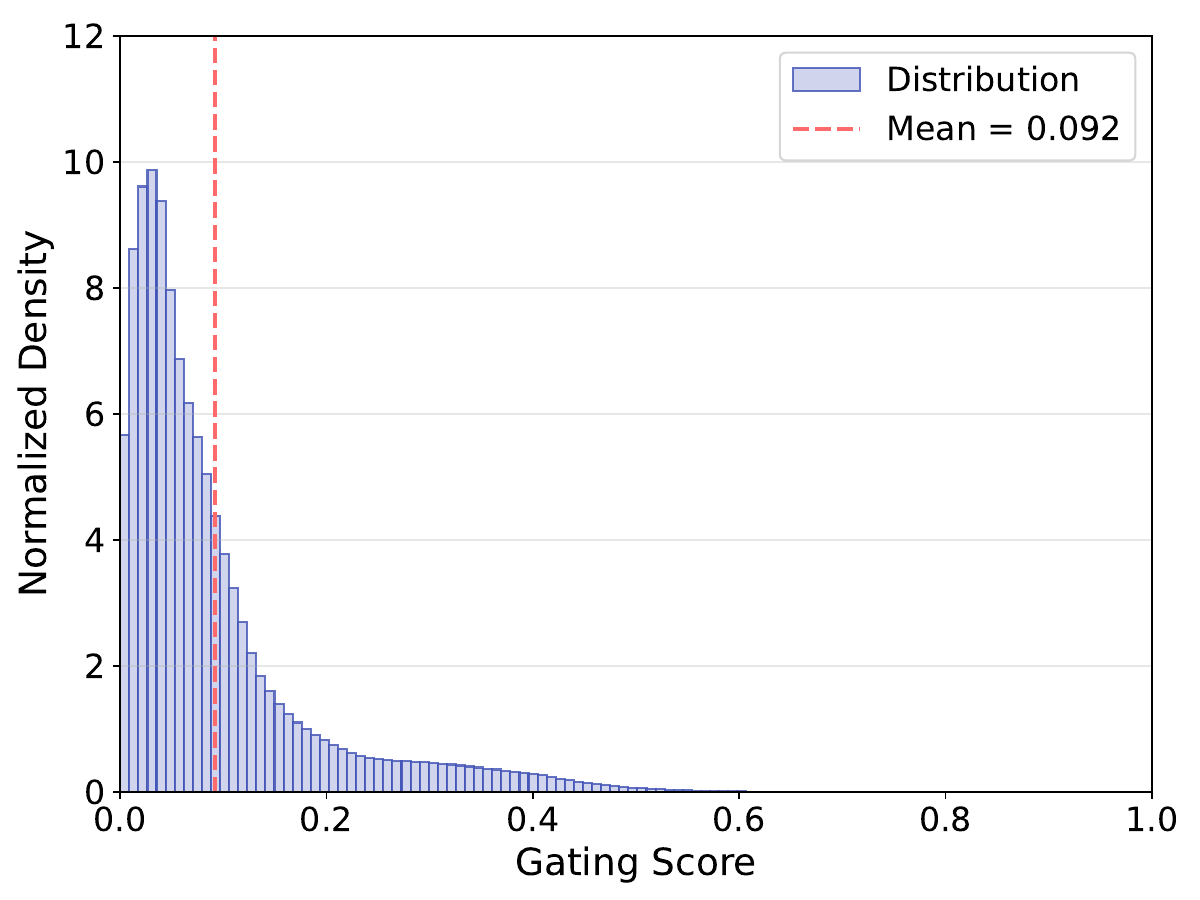}
  \end{subfigure}
  \hfill
  \begin{subfigure}[b]{0.45\textwidth}
    \label{fig:imgr_p1_box}
    \vspace{0.5em}
    \includegraphics[width=\linewidth]{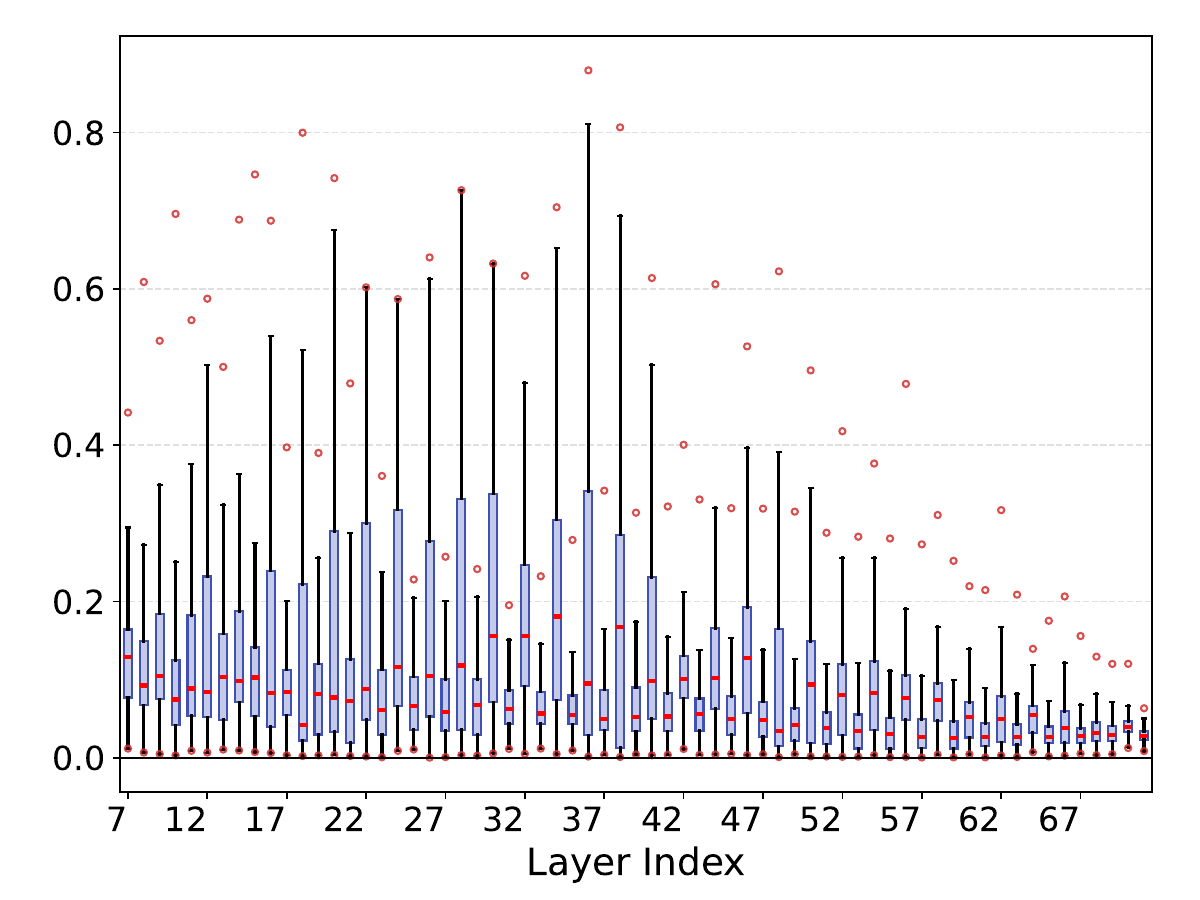}
  \end{subfigure}  
  \caption{Gating score statistics of the independent MGR ($n=8$). Distributions and means of the gate values across all layers (Left), and layerwise box plots of the gate values (Right), the red hollow circles above the box represent the maximum values (outliers) in the corresponding layers.}
  \label{fig:imgr_p1}
\end{figure}

\end{document}